\pdfoutput=1
\documentclass[11pt]{article}

\usepackage[final]{acl}

\usepackage{times}
\usepackage{latexsym}

\usepackage[T1]{fontenc}
\usepackage[utf8]{inputenc}

\usepackage{microtype}

\usepackage{inconsolata}

\usepackage{graphicx}

\usepackage{subfig}
\usepackage[commandnameprefix=ifneeded, todonotes={textsize=small}]{changes}
\definechangesauthor[color=magenta]{ga}
\definechangesauthor[color=blue]{mz}
\definechangesauthor[color=orange]{sa}
\definechangesauthor[color=red]{am}
\definechangesauthor[color=green]{ss}
\newcommand{\tabrowsep}{\addlinespace[0.1cm]}

\newcommand{\significant}{\cellcolor{red!20}}

\newcommand{\repo}{\url{https://github.com/deep-spin/gender-bias-qe-metrics}}
\usepackage{times}
\usepackage{latexsym}
\usepackage{booktabs}
\usepackage{graphicx}
\usepackage{booktabs}
\usepackage{multirow}
\usepackage{xcolor}
\usepackage{amssymb}
\usepackage{colortbl}
\usepackage{mathtools}
\usepackage{subcaption}
\usepackage{pifont}
\usepackage{fontawesome}
\usepackage{arydshln}
\usepackage{nowidow}
\usepackage{tabularx}
\usepackage{enumitem}
\usepackage{changes}
\usepackage{amssymb}% http://ctan.org/pkg/amssymb
\usepackage{pifont}% http://ctan.org/pkg/pifont
\title{Watching the Watchers:\\ Exposing Gender Disparities in Machine Translation Quality Estimation}

\newcommand{\IT}{$^{\clubsuit}$}
\newcommand{\ist}{$^{\spadesuit}$}
\newcommand{\unb}{$^{\diamondsuit}$}

\author{Emmanouil Zaranis$^*$\IT\ist, Giuseppe Attanasio$^*$\IT, Sweta Agrawal\IT, André F.T. Martins\IT \ist \unb \\
  \IT Instituto de Telecomunicações, Lisbon, Portugal \\
  \ist Instituto Superior Técnico, Universidade de Lisboa, Portugal\\
  \unb Unbabel, Lisbon, Portugal \\
\texttt{emmanouil.zaranis@tecnico.ulisboa.pt}}

\begin{document}
\maketitle
\begin{abstract}
Quality estimation (QE)---the automatic assessment of translation quality---has recently become crucial across several stages of the translation pipeline, from data curation to training and decoding. While QE metrics have been optimized to align with human judgments, whether they encode social biases has been largely overlooked. Biased QE risks favoring certain demographic groups over others, e.g., by exacerbating gaps in visibility and usability.
This paper defines and investigates gender bias of QE metrics and discusses its downstream implications for machine translation (MT).
% Focusing on out-of-English translations into languages with grammatical gender, we ask: Do contemporary QE metrics exhibit gender bias? Can the use of contextual information mitigate this bias? How does QE influence gender bias in MT outputs?
Experiments with state-of-the-art QE metrics across multiple domains, datasets, and languages reveal significant bias.
When a human entity's gender in the source is undisclosed, masculine-inflected translations
score higher than feminine-inflected ones, and gender-neutral translations are penalized.
Even when contextual cues disambiguate gender, using context-aware QE metrics leads to more errors in selecting the correct translation inflection for feminine referents than for masculine ones.
Moreover, a biased QE metric affects data filtering and quality-aware decoding.
Our findings underscore the need for a renewed focus on developing and evaluating QE metrics centered on gender.\footnote{We release code and artifacts at: \repo.}

\def\thefootnote{*}\footnotetext{Equal contribution.}\def\thefootnote{\arabic{footnote}}

\end{abstract}

\section{Introduction}
\label{sec:introduction}

\begin{figure*}[!t]
    \centering
    \includegraphics[width=0.80\linewidth]{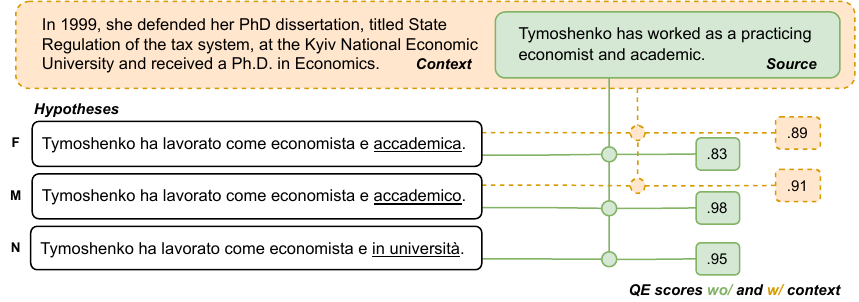}
    \caption{\textbf{Our contrastive setup.} MT-GenEval \citep{currey-etal-2022-mt} sample of an English source sentence (solid green box), its preceding context (dashed orange box), and three minimal-edit Italian translations (bottom left) evaluated by xCOMET XL \citep{guerreiro2024xcomet}. Based on the entity's (Tymoshenko) gender, the noun ``\underline{academic}'' can inflect into a feminine (F) or masculine (M) form, or be rephrased with gender-neutral (N) language.
    QE scores without context (solid) for all three forms and for the F/M (dashed) when context disambiguating gender is considered. 
    The masculine translation scores higher, regardless of whether the context is provided or not.}
    \label{fig:1}
\end{figure*}

Quality estimation---the automatic evaluation of machine-translated content without human-written references~\citep{callison-burch-etal-2012-findings}---has gained increasing interest in the natural language processing (NLP) and machine translation communities. Recent work has focused on building QE metrics aligned with human quality judgments using neural encoders \cite{rei-etal-2020-unbabels, perrella-etal-2022-matese, rei-etal-2023-scaling, juraska-etal-2023-metricx, guerreiro2024xcomet} or autoregressive language models \cite{kocmi-federmann-2023-large, fernandes-etal-2023-devil,kocmi-federmann-2023-gemba, xu-etal-2023-instructscore}, and has explored their use throughout the translation pipeline, e.g., for data filtering \citep{peter-etal-2023-theres, alves2024tower}, training \citep{ramos-etal-2024-aligning, yan-etal-2023-bleurt,he-etal-2024-improving}, and improving output quality \citep{freitag2022high, fernandes-etal-2022-quality,farinhas-etal-2023-empirical, ramos2023aligning, vernikos-popescu-belis-2024-dont}. 

Evidence has shown QE scores effectively measure translation quality aspects 
like adequacy and fluency \citep{guerreiro2024xcomet}. However, little is known about whether exogenous factors, e.g., conformity to social norms, play a role. If QE metrics favor normative language, they may penalize non-conforming expressions. Gender norms exemplify this issue---gender-based discrimination can amplify certain groups' visibility in favor of others, leading to representational and allocative harms \citep{crawford2017trouble}. 
For instance, biased metrics may systematically retain masculine-inflected data in parallel corpus filtering while discarding equivalent-quality feminine translations. Similarly, when using QE to improve MT systems \citep{fernandes-etal-2022-quality}, such metrics could lead to systems that predominantly generate masculine translations, forcing women users to spend more time and effort correcting gender errors \citep{savoldi2024harm}.

In this paper, we conceptualize and measure gender bias in QE metrics and its downstream effects.
We draw two conditions to identify a biased metric.
A QE metric exhibits gender bias if, when translating human entities, it systematically
(\textit{i}) assigns higher scores to one gender form over others when no explicit lexical cues disclose gender, or
(\textit{ii}) fails to detect the correct gender form even when lexical cues are provided, disproportionately across genders.
We investigate bias across commonly used QE metrics through an extensive empirical study. We focus on out-of-English translation where the target language uses grammatical gender,\footnote{I.e., where gender for human referents is assigned on a semantic basis and reflected with morphological marks.} a scenario frequently used in prior research on gender bias in MT \citep[e.g.,][]{vanmassenhove-etal-2018-getting,stanovsky-etal-2019-evaluating,saunders-etal-2020-neural,attanasio-etal-2023-tale,sant-etal-2024-power}. We address both binary (i.e., contrasting masculine and feminine translations) and gender-neutral (i.e., contrasting gendered and gender-neutral forms) translation scenarios, and experiment with 11 state-of-the-art metrics (supervised and prompted) across two domains, three datasets, and eight target languages.

\textbf{Our analysis reveals evidence of gender bias} across metrics, languages, and experimental conditions. Masculine forms receive higher scores when the source gender is ambiguous. Moreover, metrics favor gendered over neutral expressions. Even with disambiguating context, metrics show higher error rates in identifying feminine gender forms. These biases significantly impact MT pipelines: feminine-inflected translations fail quality thresholds more frequently than masculine equivalents despite equal validity. During inference, QE-based hypothesis reranking amplifies gender bias in MT while improving translation quality. These findings call for extending QE metrics evaluation beyond human judgment alignment to consider gender equity.

\section{Related Work}

\subsection{Evaluating MT Metrics}  
Significant efforts have been made to evaluate automatic metrics for MT. Most have focused on comparing sentence- or system-level correlations between a metric's score and human judgements \citep{machacek-bojar-2014-results,stanojevic-etal-2015-results,kocmi-etal-2021-ship,freitag-etal-2023-results,deutsch-etal-2023-ties,thompson_2024_improving_stat_sign}.
% collected through DA \citep{graham-etal-2013-continuous}. \sweta{its not just DA, its MQM, Post edits, etc}
Finer-grained analyses have discovered poor robustness to low-quality translations \citep{fomicheva-specia-2019-taking}, domain \citep{zouhar-etal-2024-fine}, or passive voice \citep{avramidis-etal-2023-challenging}. Metrics have also been shown to produce imbalanced score distributions~\citep{sun-etal-2020-estimating}, and fail to distinguish discrepancies in numbers and named entities \citep{amrhein-sennrich-2022-identifying} or high-quality translations~\citep{agrawal2024can}.
However, none of the studies have thus far evaluated gender bias exhibited by QE metrics.

\subsection{Gender Bias in MT}
Gender bias in MT systems---stemming from societal norms, model design, and deployment decisions---has been extensively documented \citep{savoldi-etal-2021-gender}. Prior work in NLP has primarily focused on measuring the extent of such bias in system outputs~\citep[e.g.,][]{stanovsky-etal-2019-evaluating,cho-etal-2019-measuring,hovy-etal-2020-sound,vanmassenhove-monti-2021-gender,levy-etal-2021-collecting-large,lucy-bamman-2021-gender,sant-etal-2024-power,piergentili-etal-2023-hi,lauscher-etal-2023-em,lardelli-etal-2024-building,stewart-mihalcea-2024-whose}, as well as on proposing mitigation strategies~\citep[e.g.,][]{escude-font-costa-jussa-2019-equalizing,costa-jussa-de-jorge-2020-fine,saunders-etal-2020-neural,saunders-byrne-2020-reducing,escolano-etal-2021-multi,lee-etal-2023-target,garg-etal-2024-generating}. In contrast to this line of work, we shift the focus from the translation systems themselves to the task of quality estimation, a step that, typically, follows the MT. Further, we investigate whether  bias in QE can propagate to or reinforce bias in MT outputs.

\subsection{Social Biases in NLG Metrics}
Several studies have investigated whether metrics used to evaluate natural language generation (NLG) exhibit social biases.
% In contrastive setups similar to ours, 
\citet{qiu-etal-2023-gender} test whether \textit{n}-gram- and model-based metrics, \textit{e.g.} CLIPScore~\citep{hessel-etal-2021-clipscore} show gender bias in image captioning.
\citet{sun-etal-2022-bertscore} quantify several social biases. They use English coreference resolution examples to measure the gender bias of several \textit{n}-grams and model-based NLG metrics, including BERTScore~\citep{bert-score}, MoverScore~\citep{zhao-etal-2019-moverscore}, and BLEURT~\citep{sellam-etal-2020-bleurt}.
\citet{gao_social_biases_automatic_eval_metrics_nlg} measure race and gender stereotypes using WEAT \citep{caliskan2017semantics} and SEAT \citep{may-etal-2019-measuring}. 
However, no study targets QE metrics for MT or reference-free NLG metrics.

\section{Measuring Gender Bias in QE}

Several design choices arise when testing how neural QE metrics handle gender-related translations, including
(\textit{i}) source and target languages, (\textit{ii}) domain,
(\textit{iii}) granularity (document, passage, or sentence-level MT), (\textit{iv}) gender ambiguity in the source, and (\textit{v}) 
% words to monitor for gender-inflection 
gender-inflected words in the target.
In this work, we experiment with \textbf{out-of-English} (En$\rightarrow$*) MT setups where the target language (*) is characterized by grammatical gender. In these languages, masculine and feminine genders are represented through word choices and inflection (e.g., different suffixes, as in Figure~\ref{fig:1}), while neo-pronouns, neo-morphemes, or rephrasing permit beyond-the-binary or neutral translation \citep{lauscher-etal-2023-em,piergentili-etal-2023-hi,lardelli-etal-2024-building,piergentili-etal-2024-enhancing}.
Following prominent work on gender bias in MT \citep{currey-etal-2022-mt,Rarrick2023GATEAC}, we focus on \textbf{sentence-level} translations from \textbf{multiple domains}
% ---a setup dominant in the literature \citep{post2023escaping}---
and observe gender-inflected words referring to \textbf{human entities}.
% , also known as Gender-Marked Entities \citep[GME;][]{Rarrick2023GATEAC}.
We challenge QE metrics with translation pairs where the entity's gender in the source is ambiguous or contextual cues disambiguate it.

\begin{figure*}[!t]
    \centering
    \includegraphics[width=0.8\linewidth]{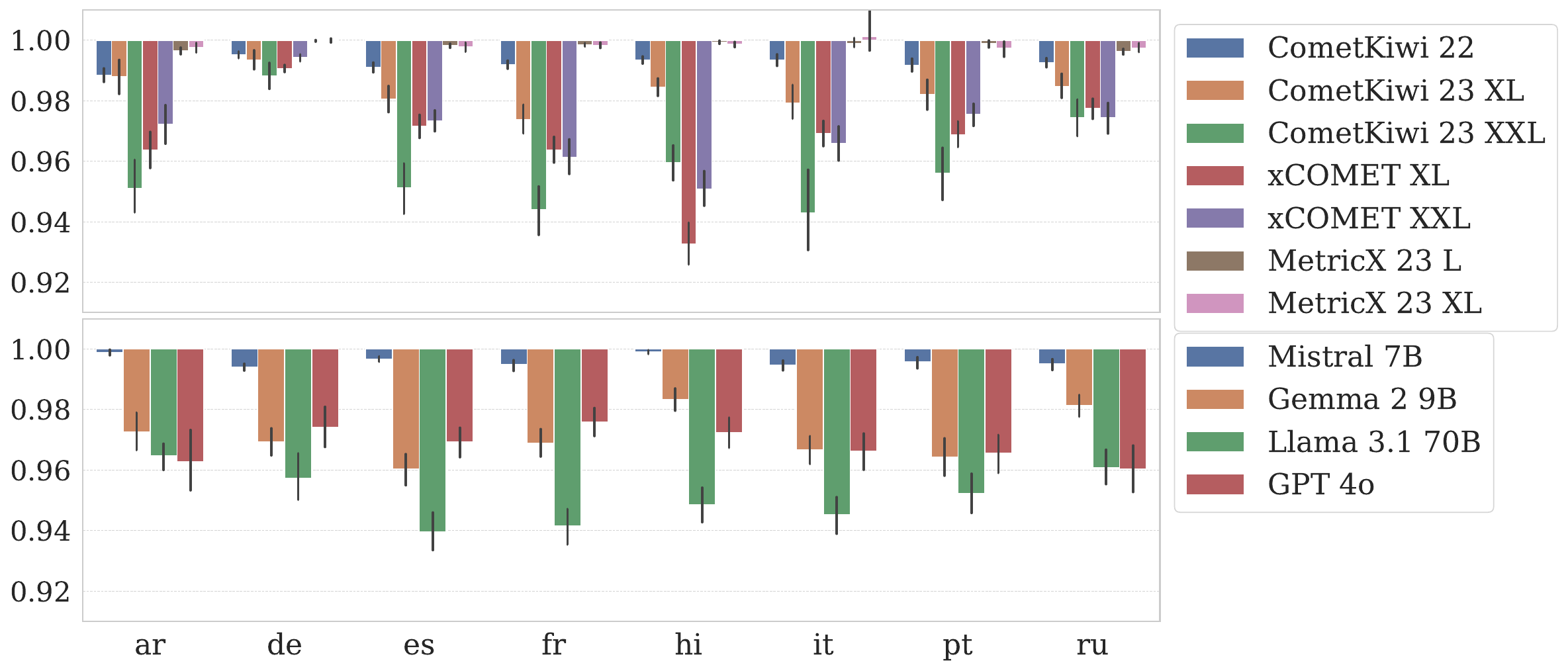}
    \caption{\textbf{Ratio $\mathrm{QE}(s,h_F) /\ \mathrm{QE}(s,h_M)$ on MT-GenEval, ambiguous instances}. Average and 95\% confidence interval on the test set by language. Neural (top) and GEMBA (bottom) QE metrics.}
    \label{fig:ratio_MTGenEval_ambiguous}
\end{figure*}

\subsection{Formalizing Quality Estimation}\label{subsec:formalizing_qe}
For the remainder of the paper, we will discuss QE metrics that support quality assessment of translations \cite{graham-etal-2016-glitters,burchardt-2013-multidimensional}. A QE metric is thus a function $f$ such that:
\begin{equation}
    \mathrm{QE}(s, h) \coloneqq f_\theta(s, h) = a,
\end{equation}
where 
% ``$;$'' denotes simple string concatenation,
$s$ is the English source, $h$ the translation to evaluate, 
% $c$ is an optional context for \textit{s},
and $a \in \mathbb{R}$ is the assessment score. Typically, $f$ is a neural function parameterized by $\theta$, and \textit{a} is bounded in a range with the edges indicating the poorest and highest quality. 

\paragraph{QE Metrics} We test 11 QE metrics, varying model sizes and architectures. We include seven neural metrics that ranked highest in recent editions of the WMT QE Shared Task \citep{blain-etal-2023-findings}: CometKiwi 22 \citep{rei-etal-2022-cometkiwi}, CometKiwi 23 XL/XXL \citep{rei-etal-2023-scaling}, xCOMET XL/XXL \citep{guerreiro2024xcomet}, and MetricX 23 L/XL \citep{juraska-etal-2023-metricx}.
Following \citet{kocmi-federmann-2023-large}, we test the GEMBA-DA prompt with three decoder-only, instruction-tuned, open-weight language models---Mistral 7B \citep{jiang2023mistral}, Gemma 2 9B \citep{team2024gemma}, Llama 3.1 70B \citep{dubey2024llama}---and a commercial API model, GPT 4o \citep{achiam2023gpt}.

Our analysis will investigate whether metrics can use extra-sentential contextual cues (\S\ref{subsec:gender_unambiguous_entities}).
However, contemporary QE metrics are not trained on nor support context beyond the sentence level off-the-shelf.
Thus, we rely on two inference-time strategies to incorporate
% experiment with
extra-sentential context \cite{vernikos-etal-2022-embarrassingly}:
% Given the context $c$ that disambiguates the gender identity of the referent in the source, 
% We experiment with two variants: 
1) We concatenate the context string $c$ to $s$ and $h$, i.e., $\mathrm{QE}(c;s, c;h)$, and 2) we apply the same strategy as (1) 
% for the source 
but automatically translate the
% the 
context $c$ for the hypothesis, 
% side into the target language beforehand
i.e., $ \mathrm{QE}(c;s, \mathrm{translate}(c);h)$.\footnote{We use NLLB-200 3.3B \citep{NLLB_arxiv} for translating $c$ as it supports all studied languages.} 
The latter variant prevents the creation of a cross-lingual hypothesis, out-of-distribution for the QE metric. We modify the prompt for GEMBA metrics to account for context (full details in Appendix~\ref{ssec:app:computing_DA_scores}).

\subsection{Experimental Conditions}

In \S\ref{sec:introduction}, we defined a biased QE metric as one that exhibits systematic undue preferences for gender forms in the target or disproportionate error rate across gender groups. 
Here, we describe the datasets and measurement methods we use to assess the presence and extent of gender bias. % this phenomenon.

\subsubsection{Datasets with Minimal Edits}
\label{sssec:datasets}

We study datasets organized in \textbf{minimal edit} contrastive pairs, where sources and \textbf{reference translations} differ only in gender-related words. This approach has two main advantages. First, it enables precise isolation of effects related to gender-inflected elements (see, for example, translations in Figure~\ref{fig:1}). Therefore, any change in the QE score can only be attributed to the metric's sensitivity to those words.
Second, using human-written references neutralizes potential confounding quality-related effects.\footnote{Sentence-level outputs are converging toward human-level translation quality in MT research; thus, using references is a valid choice. However, we relax this constraint in \S\ref{subsec:findings_GT_translations}.}
% Minimal edits permit to isolate the effect of gender. 

\subsubsection{Gender-Ambiguous Entities}

With gender ambiguity in the source, we set out to study two contrastive setups: masculine (m) vs. feminine (f) and gendered vs. gender-neutral translations. For the former (m/f) experiments, we use MT-GenEval \citep{currey-etal-2022-mt} and GATE \citep{Rarrick2023GATEAC}, two established sentence-level corpora for gender bias evaluation in MT with translations provided by professional translators. From MT-GenEval, we retrieve all test examples from the \textit{contextual} subset, discarding the context that would disambiguate the gender. From GATE, we collect all instances with one gender-marked entity. The two corpora provide linguistic diversity: MT-GenEval collects sentences from Wikipedia, while  %, i.e., the encyclopedic domain.
sources in GATE are written by linguists and cover diverse linguistic phenomena.
% for the GME and length
% (see Appendix A in \citet{Rarrick2023GATEAC}). 
MT-GenEval includes eight target languages (ar, de, es, fr, hi, it, pt, ru) and GATE three (es, fr, it). 
% Crucially, references in both datasets come in \textbf{minimal edit pairs}, i.e., they differ only for the gender-marked words (see F/M hypotheses in Figure~\ref{fig:1}).
Following prior work \citep{beutel2019putting, gaut-etal-2020-towards, Attanasio2024TwistsHA}, we observe gaps in relativistic terms. We measure the ratio $\mathrm{QE}(s,h_F) /\ \mathrm{QE}(s, h_M)$ where $h_F$ and $h_M$ are the feminine- and masculine-inflected translations, respectively. 
Echoing the desiderata of counterfactual fairness \citep{czarnowska-etal-2021-quantifying}, the two scores should not differ, i.e., the ideal ratio is 1.

We use mGeNTE \citep{savoldi2025mgentemultilingualresourcegenderneutral} for gender-neutral experiments. The benchmark targets gender-neutral translations of English sources written by professional translators for it, es, de.\footnote{Neutralization strategies might extend beyond simple edits. While breaking the minimal-edits requirement, these instances highlight meaningful contrasts between common gendered translations and less common neutral ones.}
We focus on ambiguous sources (Set-N), and measure $\mathrm{QE}(s,h_N) /\ \mathrm{QE}(s, h_G)$, the ratio between the gender-neutral and gendered score. Since gender-neutral translations provide inclusive alternatives that maintain semantic accuracy, QE metrics should assign them higher scores than their gendered counterparts, i.e., the ratio should ideally exceed 1. However, equality between scores represents a minimum acceptable threshold.

\subsubsection{Gender-Unambiguous Entities}\label{subsec:gender_unambiguous_entities}

We study cases where intra- or extra-sentential lexical cues disambiguate gender.
For the intra-sentential case, we use the MT-GenEval's \textit{counterfactual} subset. It consists of paired instances with masculine- and feminine-inflected sources, each with its corresponding reference. For the extra-sentential case, we use the MT-GenEval's \textit{contextual} subset, where each test instance is paired with a preceding sentence disambiguating the gender. Figure~\ref{fig:1} reports an example of the extra-sentential case.
% cues in the preceding sentence. 
An instance with intra-sentential cues would be: ``In \underline{her} thirties, Tymoshenko has worked as a practicing economist and academic.''

Since a single correct gender form exists, given a set of instances $S$, we compute the error rate ($\mathrm{ER}$)---the fraction of instances where the QE metric assigns a higher score to the incorrect gender form. 
We compute it for all instances and separately for feminine and masculine sources, i.e., $\mathrm{ER}(S^{F})$ and $\mathrm{ER}(S^{M})$, respectively.\footnote{We count ties as errors since they indicate the system's failure to select the correct gender inflection.} 
To measure gender disparities, we adopt a Multi-group Comparison Metric \citep{czarnowska-etal-2021-quantifying} defined as:
\begin{equation}
    \Phi(S^{F}, S^{M}) = \mathrm{ER}(S^{F}) /\ \mathrm{ER}(S^{M}) 
\end{equation}

Loosely inspired by False Positive Equality Difference \citep{Dixon2018MeasuringAM}, the metric operationalizes the statement: ``Even when lexical cues leave no doubts on gender identity, sources with feminine referents exhibit $\Phi$ times the number of errors in gender inflection as sources with masculine referents.'' The ideal scenario is equal errors, i.e., a ratio of 1. We report $QE(s_F, h_F)/QE(s_M, h_M)$, i.e., the instance-level ratio of the QE scores assigned to feminine and masculine pairs, in Appendix~\ref{app:ssec:results_gender_unambiguous}. 

\section{Findings}  \label{sec:findings}

This section presents the results for ambiguous (\S\ref{ssec: ambiguous}), unambiguous with intra-sentential (\S\ref{subsec:intra_sentencial}) and extra-sentential cues (\S\ref{subsec:extra_sentential}) pairs.
Our analysis reveals significant ($p<.05$) gender bias across most evaluated metrics and configurations (see Appendix~\ref{ssec:statistical_significance} for details on statistical testing).

\begin{table}[!t]
\centering
\small
\begin{tabular}{lccc}
\toprule
\textbf{Metric} & \textbf{$\mathrm{ER}$} & \textbf{$\Phi$}  \\
\midrule
CometKiwi 22  & 0.11 & 1.70 \\
CometKiwi 23 XL & 0.09 & 1.18  \\
CometKiwi 23 XXL & \underline{\textbf{0.07}} & \underline{\textbf{0.87}}\\
xCOMET XL & 0.10 &  1.81 \\
xCOMET XXL & 0.08& 1.32 \\
MetricX 23 L & 0.31 & 1.25 \\
MetricX 23 XL  & 0.12 & 1.19 \\
\midrule
Llama 3.1 70B & 0.31 & 1.16   \\
Gemma 2 9B & 0.28 & 1.36  \\
Mistral 7B & 0.74 & \textbf{1.13}  \\
GPT 4o & \textbf{0.16} & 1.15  \\
\bottomrule
\end{tabular}
\caption{\textbf{Total error rate ($\mathrm{ER} \downarrow$) and error rate ratio between gender groups ($\Phi(S^F,S^M) \rightarrow1$)}. MT-GenEval \textit{counterfactual} set, unambiguous (intra-sentential). Best metrics per type in bold, best overall, underlined. Mean results across eight languages.}
\label{tab:results_counterfactual_errors}
\end{table}

\subsection{Gender-Ambiguous Instances} \label{ssec: ambiguous}
\label{ssec:ambiguous_case}

\textbf{Most QE metrics and language pairs show a significantly higher score for masculine-inflected references.}
Figure~\ref{fig:ratio_MTGenEval_ambiguous} reports the average f/m ratio of QE scores on MT-GenEval test sets (full raw scores are in Figure~\ref{fig:raw_scores_MTGenEval_contextual_ambiguous} of Appendix~\ref{app:ssec:results_gender_ambiguous}).
We found no significant difference between f/m scores only for MetricX (X and XL), en$\rightarrow$de. The metric is generally the closest to parity across languages. 
We hypothesize that MetricX models, unlike Comet, are trained on additional synthetic data that attributes perfect scores to reference translations. This choice likely skews the metric results 
by making it more sensitive to identifying references than quality differences from minimal edits.
Among the best models, Mistral 7B reaches an average ratio of 0.9961. Notably, preference toward the masculine-inflected translations increases with size for both CometKiwi (22 < 23 XL < 23 XXL) and GEMBA models (Mistral 7B < Gemma 2 9B < Llama 3.1 70B). 
These results hold for GATE (see Figures~\ref{fig:raw_scores_GATE} and ~\ref{fig:ratio_GATE} in Appendix~\ref{app:ssec:results_gender_ambiguous}).

\textbf{Most QE metrics score gender-neutral translations consistently lower than gendered ones on all languages.}
Figure~\ref{fig:ratio_mGeNTE} in Appendix~\ref{app:ssec:results_gender_ambiguous} reports the ratio between QE scores for gendered and gender-neutral translations on mGeNTE's ambiguous instances.
Despite the unknown gender in English, all metrics but MetricX and Mistral 7B prefer 
% significantly 
gendered translations significantly. Similar to the f/m setup, we observe that the larger a CometKiwi variant is, so is the 
% stronger the 
preference for gendered translations. Findings are consistent across languages, with a larger magnitude on en$\rightarrow$es, suggesting that QE metrics may still be unsuitable for assessing gender-fair forms \cite{sczesny2016can}.

\subsection{Gender-Unambiguous Instances}
\label{ssec:gender_unambiguous}
Tables~\ref{tab:results_counterfactual_errors} and ~\ref{tab:results_contextual_non_amb_aggr} present the total error rate ($\mathrm{ER}$) and error rate ratio ($\Phi(S^F, S^M)$) 
% for all metric variants 
for the intra and extra-sentential cases across metrics.

\subsubsection{Intra-sentential Cues}
\label{subsec:intra_sentencial}

\textbf{Most metrics consistently exhibit a higher frequency of errors when the source mentions a feminine entity than a masculine one.} As shown in Table~\ref{tab:results_counterfactual_errors}, the majority of metrics yield values greater than 1 for $\Phi(S^F, S^M)$, indicating increased susceptibility to errors when tasked to select the correctly inflected translation and the referent in the source is feminine. CometKiwi 23 XXL is the only exception, yielding the least biased behavior overall; however, it still exhibits bias in 2 out of 8 languages (full results disaggregated by language are in Table~\ref{app:tab:counterfactual_stat_tests_ratio_Phi} of Appendix~\ref{app:ssec:results_gender_unambiguous}).

\subsubsection{Extra-sentential Cues}
\label{subsec:extra_sentential}

\begin{table}[t]
\centering
\small
% \resizebox{.99\linewidth}{!}{%
\begin{tabular}{lcccc}
\toprule
\textbf{Metric} & \multicolumn{2}{c}{} & \multicolumn{2}{c}{$\mathrm{tr}(c)=\checkmark$}\\
& \textbf{$\mathrm{ER}$} & \textbf{$\Phi$} &\textbf{$\mathrm{ER}$} & \textbf{$\Phi$}   \\
\midrule
CometKiwi 22  & 0.28 & 0.84  & 0.19 & 6.68  \\
CometKiwi 23 XL   & 0.28 & \underline{\textbf{0.92}} & 0.17 & 5.09  \\
CometKiwi 23 XXL  & \underline{\textbf{0.14}} & 2.20   & \underline{\textbf{0.14}} & 4.85  \\
xCOMET XL  & 0.29 & 4.53 & 0.24 & 7.03  \\
xCOMET XXL  & 0.20 & 4.97  & 0.20 & 5.47  \\
MetricX 23 L  & 0.39 & 0.73 & 0.34 & 1.53  \\
MetricX 23 XL & 0.22 & 1.54 & 0.28 & 1.60 \\
\midrule
Llama 3.1 70B  & 0.61 & 2.20 & - & -  \\
Gemma 2 9B  & 0.57 & 2.71 & - & -  \\
Mistral 7B  & 0.88 & \textbf{1.22} & - & -  \\
GPT 4o & \textbf{0.44} & 2.64 & - & -  \\
\bottomrule
\end{tabular}%
% }
\caption{\textbf{Total error rate ($\mathrm{ER} \downarrow$) and error rate ratio between groups ($\Phi(S^F,S^M) \rightarrow1$)}. MT-GenEval \textit{contextual} set, unambiguous (extra-sentential). $\mathrm{tr}(c)=\checkmark$: context-aware metrics with translated context. Best models per type in bold, best overall, underlined. Mean results across eight languages.}
\label{tab:results_contextual_non_amb_aggr}
\end{table}

\begin{figure*}[!t]
    \centering
    \includegraphics[width=0.70\linewidth]{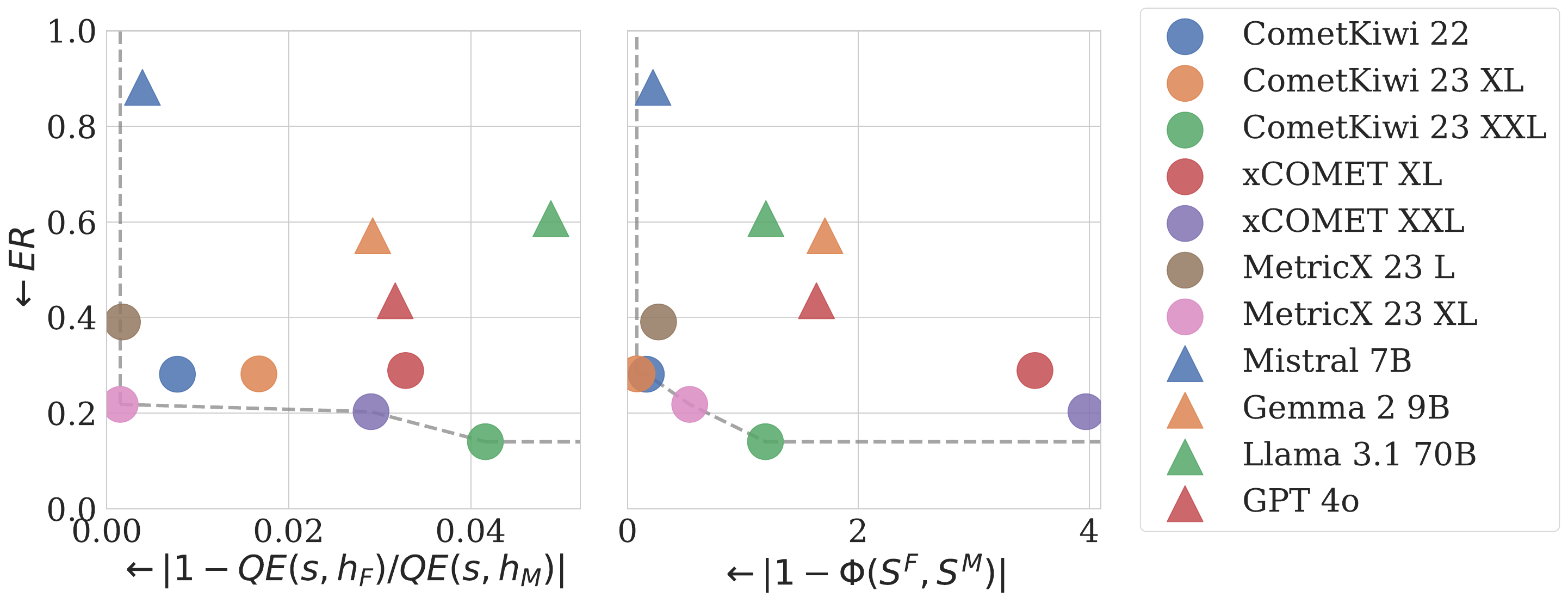}
    \caption{\textbf{Total error rate vs. Gaps from parity.} X-axes show gaps from parity for ambiguous (left) and unambiguous (right) instances. Dashed lines indicate the Pareto frontier. Ideally, a metric should achieve both a low error rate and parity, resulting in proximity to the origin (0,0). Neural metrics use the original context.
    % We use the MON context-aware versions of the metrics from Table~\ref{tab:results_contextual_non_amb_aggr}.
    }
\label{fig:amb_vs_non_amb.pdf}
\end{figure*}

\textbf{Most QE metrics exhibit a significantly higher error rate on feminine sources than masculine sources.}
When the context is not translated, four out of seven context-aware neural metric variants make at least 1.5 times as many errors on feminine sources as, on average, masculine ones. CometKiwi 23 XL yields the best error rate ratio among neural metrics, followed by CometKiwi 22 and MetricX 23 L.

Metrics that use translated context yield lower overall error rates. However, improved accuracy is primarily driven by better results on masculine sources. Therefore, \textbf{translated context exacerbates gender disparities for all QE metrics} (see full results in Table~\ref{app:tab:results_contextual_non_amb_aggr} of Appendix~\ref{app:ssec:results_gender_unambiguous}).
$\Phi(S^F, S^M)$ increases on average 3 times, with the highest increase accounting for CometKiwi 22 (approximately 8 times higher error rate).\footnote{We observed similar trends even when using Google Translate for translating $c$ (see Appendix~\ref{app:ssec:results_gender_unambiguous}).}
We hypothesize that this significant increase might be attributed to the translated context, which may induce bias towards masculine referents.
These results suggest that only looking into metrics optimizing the overall error rate can be misleading and result in higher disparities.

Regarding GEMBA metrics, only Mistral 7B shows a balanced error rate between genders. On the contrary, Llama 3.1, Gemma 2, and GPT-4o make approximately two to three times as many errors on feminine sources as on masculine sources. These findings indicate a strong preference for masculine-inflected translations, even when gender cues are available in the preceding context. 

However, upon a closer inspection, we noted that \textbf{GEMBA QE metrics cannot distinguish gender inflections successfully.}
Ties---i.e., an equal DA score for the correct and incorrect translation---are widespread. On average, Mistral 7B assigns the same score to 85\%, Gemma 2 9B to 50\%, Llama 3.1 70B to 55\%, and GPT-4o to 38\% of test samples (full details in Table~\ref{app:tab:ties_contextual_non_amb_llms}, Appendix~\ref{app:ssec:results_gender_unambiguous}). Even with fewer ties, (Gemma 2 and GPT-4o) metrics show a significantly ($p <.05$) higher error rate for feminine sources. See Table~\ref{app:tab:contextual_non_amb_stat_tests_phi} of Appendix~\ref{app:ssec:results_gender_unambiguous} for results on individual languages. We explore further GEMBA models' brittleness to surface-level differences in the following section.

\subsection{Bias and Surface-Level Brittleness} \label{ssec:overall}

An unbiased QE metric evaluates masculine and feminine translations equivalently when the source gender is unknown.
However, whether such scores derive from an unbiased assessment or failure to detect gender markers remains unresolved, which could compromise accuracy and usability when gender is known.
To investigate this distinction, we analyze the metrics results jointly across ambiguous and unambiguous scenarios.

Figure~\ref{fig:amb_vs_non_amb.pdf} reports the total error rate 
% ($ER\downarrow$) 
when source gender is known---a measure of overall accuracy---vs. gender bias measurements in ambiguous ($\mathrm{QE}(s,h_F) / \mathrm{QE}(s,h_M)$) and unambiguous cases ($\Phi(S^F, S^M)$). For simplicity, we report one minus these values to have the optimal result at 0. The chart reveals that among the metrics identified as least biased metrics in the ambiguous case (\S\ref{ssec:ambiguous_case}), 
only MetricX 23 XL can distinguish gender inflections in the unambiguous case ($\mathrm{ER}<.25$). In contrast,
Mistral 7B scores many ties (Table~\ref{app:tab:ties_contextual_non_amb_llms} in Appendix~\ref{app:ssec:results_gender_unambiguous}), demonstrating a low bias in both setups but an astonishingly high error rate (and extreme brittleness to gender-related surface-level differences). 
More generally, this shortcoming can be attributed to the tendency of GEMBA metrics to assign coarse-grained scores \cite{zheng_2023_llms_multiple_choice_selectors,stureborg_llms_biased_eval_2024}---typically 85, 90, 95, and 100 in our experiments. 
Based on this result, we discourage using GEMBA-prompted language models if capturing gender-related phenomena is crucial for the use case at hand.

CometKiwi 23 XXL is a better alternative, being the only other Pareto-optimal metric in ambiguous and unambiguous setups with a low overall error rate. 
Comparing the xCOMET and Kiwi families, we note that they achieve similar gender error rates, but the latter is fairer, on average.
This holistic view is critical, as it can inform the selection of metrics for downstream applications where accurate and unbiased evaluations are necessary. 

\section{Downstream Implications}\label{sec:downstream_implications}

Previous sections have collected evidence of widespread gender bias in contemporary QE metrics.
This section explores the downstream implications of using biased QE metrics in different stages of a classical MT pipeline.

\begin{figure}[!t]
    \centering
    \includegraphics[width=0.9\linewidth]{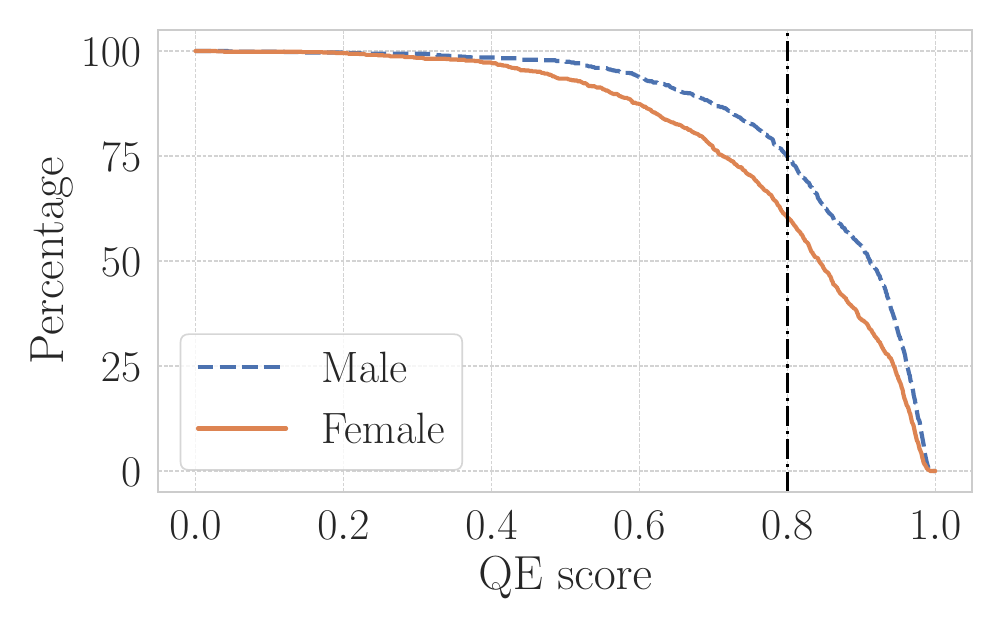}
    \caption{\textbf{Share of instances (y-axis, \%) scoring at least the value on the x-axis.} Vertical line indicates a typical data filtering threshold. MT-GenEval (ambiguous), En$\rightarrow$It, CometKiwi 23 XXL. }
    \label{fig:ccdf_ambiguous}
\end{figure}

\subsection{Quality Filtering}
\label{ssec:quality_filtering}

Filtering high-quality parallel data from web crawls is a long-standing challenge in MT research. 
Evolving from traditional approaches using heuristics \citep{resnik-smith-2003-web} or noise detection techniques \citep{taghipour-etal-2011-parallel}, recent work has adopted QE metrics to filter large corpora \citep{peter-etal-2023-theres,alves2024tower}. Arguably, data retained post filtering influences each subsequent step in the MT pipeline.

In \S\ref{ssec:ambiguous_case}, we show that in the absence of clear gender cues in the source, contemporary QE metrics tend to favor masculine-inflected hypotheses, casting shadows over their use as quality filters. To exemplify the issue, Figure~\ref{fig:ccdf_ambiguous} shows a complementary cumulative distribution function of a distribution of CometKiwi 23 XXL scores \citep{koenecke2020racial,Attanasio2024TwistsHA}.
The chart illustrates explicitly the number of instances that would be filtered out at a given quality threshold, separately by gender. A threshold of 0.8 would retain 75\% of masculine instances but only 63\% of feminine ones. 
This disparity highlights that even when quality is comparable, sentences with feminine forms are more likely to be filtered out, potentially amplifying gender representation imbalances.

\subsection{Machine Translation Quality Assessment}\label{subsec:findings_GT_translations}

\begin{table}[!t]
\centering
\small
\begin{tabular}{lcc}
\toprule
\textbf{Metric} & \textbf{$\mathrm{ER}_{GT}$} & \textbf{$\Phi_{GT}(S^F,S^M)$}   \\
\midrule
CometKiwi 22  & 0.13 & 1.34 \\
CometKiwi 23 XL & 0.11 & 1.60 \\
CometKiwi 23 XXL & \textbf{0.09} & 1.62 \\
xCOMET XL & 0.12 &  1.96 \\
xCOMET XXL & 0.10 & 1.23 \\
MetricX 23 L & 0.34 & \textbf{1.17} \\
MetricX 23 XL  & 0.14 & 1.33\\
\bottomrule
\end{tabular}
\caption{\textbf{Total error rate ($\mathrm{ER_{GT}} \downarrow$) and error rate ratio between gender groups ($\Phi_{GT}(S^F,S^M) \rightarrow1$)} when assessing \textit{GT translations}. MTGenEval's counterfactual (Unambiguous, Intra-sentential) subset. 
% case. Source instances of MTGenEval's counterfactual subset. 
Best results in bold. Mean results across eight languages.}
\label{tab:results_counterfactual_errors_GT}
\end{table}

Our evaluation in \S\ref{sec:findings} focuses on a controlled setup with minimally contrastive reference translations to isolate the impact of gender-inflected linguistic elements on QE metrics. However, QE metrics are primarily used to assess machine-generated translations---which often vary in style and quality (e.g., exhibiting characteristics of \textit{translationese}). We extend our analysis to investigate whether these metrics demonstrate systematic biases when applied to automatic translations. We focus on cases where intra-sentential cues resolve the gender identity of the referent, as in \S\ref{subsec:intra_sentencial}. Specifically, we translate the source-feminine ($s_F$) and source-masculine ($s_M$) texts from the counterfactual subset of MTGenEval using Google Translate (GT).\footnote{Prior work has shown that GT is less biased when the context provides sufficient information to disambiguate the gender \cite{rescigno2023gender, piazzolla2023good}.} To account for quality variations unrelated to gendered forms, we only retain the counterfactual pairs ($s_F, \mathrm{GT_F}$) and ($s_M, \mathrm{GT_M}$) where the translations $\mathrm{GT_F}$ and $\mathrm{GT_M}$ exhibit the correct gender inflection, as categorized by \citet{currey-etal-2022-mt}, and are of similar quality as measured by BLEU~\cite{papineni-etal-2002-bleu}.\footnote{Details about the filtering process are provided in Appendix~\ref{app:sec:GT_filtering_details}. Notably, approximately half of the initial instances are retained, as reported in Table~\ref{tab:filter_stats} of Appendix~\ref{app:sec:GT_filtering_details}.} Our findings reveal that \textbf{QE metrics exhibit a significant bias when assessing automatic translations}. When challenged to select the correctly inflected automatic translation, they are more prone to errors on sources with feminine referents than on those with masculine referents (Table~\ref{tab:results_counterfactual_errors_GT}).\footnote{Individual language results are in Table~\ref{app:tab:counterfactual_stat_tests_ratio_Phi_GT} (App~\ref{app:ssec:results_gender_unambiguous}).} These results are in line with those we obtained from human-written references.

\begin{table}
\centering
\small
 % \resizebox{\linewidth}{!}{
\begin{tabular}{llccc}
\toprule
\rowcolor{white}  
& & \multicolumn{2}{c}{\textbf{Comet-22}}  &  \multicolumn{1}{c}{\textbf{Match}} \\
      \multirow{-3}{*}{\textbf{Decoding}}  &   \multirow{-3}{*}{\textbf{Metric}} & $h_F\uparrow$ & $h_{M}\uparrow$ & $\delta_M \sim 0$ \\
     
\midrule
% changed decision rule
Greedy & - & 86.73 & 88.45  & -45.67  \\
QAD  & CK 22 &  \bf{87.66} & \bf{89.39}   &  -46.84 \\
 & CK 23 XXL & 87.46 & 89.11   &  \bf{-43.47} \\
 
\bottomrule
\end{tabular}
% }
\caption{\textbf{Quality and gender bias results translating with TowerInstruct-v0.2.} MT-GenEval contextual (ambiguous). Greedy and QAD decoding. Mean scores across six languages. QAD is run with two variants of CometKiwi (CK). Best values are in bold.
} 
\label{tab:reranking}
\end{table}

\subsection{Quality-Aware Decoding}
\label{ssec:qad}

So far, we conceptualized and measured gender bias as an undue preference expressed by quality metrics. However, when used to guide the translation for improved quality, a biased metric can impact and potentially amplify gender bias of MT outputs. 
To investigate this intuition, we experiment with quality-aware decoding \cite[QAD]{fernandes-etal-2022-quality}, a common downstream application of QE metrics.
Unlike greedy decoding, which selects the most likely next token at each step, QAD approaches rank N-best translation hypotheses generated by an MT model using automatic metrics.

\paragraph{Setup} As the target translation system, we experiment with TowerInstruct-v0.2~\cite{alves2024tower}, a state-of-the-art model for MT. Using greedy and QAD, we generate translations of the MT-GenEval contextual (ambiguous) sources for six supported languages: de, es, fr, it, ru, and pt. For QAD, we generate 50 hypotheses using epsilon sampling ($eps=0.02$) and perform reranking using CometKiwi 22 and CometKiwi 23 XXL.
Based on the findings in \S\ref{ssec:overall}, CometKiwi 23 XXL achieves Pareto-optimal performance 
% across both ambiguity scenarios (see Figure~\ref{fig:amb_vs_non_amb.pdf}) 
and CometKiwi 22 exhibits bias in the ambiguous case, allowing us to compare the impact of differently biased metrics.

We evaluate translation quality using reference-based \textsc{Comet-22} against masculine (\(h_M\)) and feminine (\(h_F\)) references. 
Relying on notions linking gender bias and uneven visibility \citep{savoldi-etal-2021-gender}, we measure bias in the MT system as the frequency each gender is represented in the generated hypothesis. Following \citet{currey-etal-2022-mt}, we check if specific gender-marked words, uniquely present in either \(h_M\) or \(h_F\), are present in the generated hypotheses.
% match those in either \(h_M\) or \(h_F\)
We report $\delta_M$, the difference between the number of feminine and masculine reference matches.
A $\delta_M$ close to zero indicates minimal gender bias as translations are equally matched to female and male references, while a large positive (negative) $\delta_M$ suggests that the system is biased towards female (masculine) inflections. 

\paragraph{Findings} Table~\ref{tab:reranking} presents the translation quality of outputs using both greedy and QAD methods. 
Using greedy decoding, TowerInstruct-v0.2 produces more masculine-inflected translations of higher quality,
% and of higher quality for masculine inflection   exhibits a strong bias toward generating masculine inflections,
as evidenced by absolute higher \textsc{Comet-22} scores for masculine outputs (\(h_M\)) and a large negative $\delta_M$ (-45.7).
While reranking improves overall translation quality over greedy, as measured by \textsc{Comet-22}, for both gender groups, its effect on gender disparity varies depending on the metric used. Specifically, QAD with the CometKiwi 23 XXL model enhances fairness, achieving the best absolute $\delta_M$ values for ambiguous source texts. In contrast, using a biased metric such as CometKiwi 22, further exacerbates gender disparity, as indicated by the more negative $\delta_M$ values. Therefore, careful selection and calibration of QE metrics are crucial to ensure that improvements in overall translation quality do not inadvertently amplify gender bias.

\section{Conclusion}

We formalized and measured gender bias in quality estimation metrics commonly used to assess translation quality. 
We studied translation from English into languages with grammatical gender, isolating the gender lexical phenomenon via minimal contrastive pairs. Our findings revealed concerning patterns. Through extensive experiments across languages, domains, and datasets, we found most QE metrics systematically penalize feminine forms in the target language---both when gender is ambiguous and when it is contextually clear. Similarly, gendered forms are mostly preferred over gender-neutral forms.  
Moreover, we show that QE bias can affect data quality filtering and quality-aware decoding with MT systems.

We should, indeed, watch the watchers. Undue gendered preferences call for a new focus on how we develop and evaluate QE metrics. Future work should center gender representativeness in training data for QE metrics, including and normalizing gender-neutral instances. Moreover, new scrutiny of QE metrics entails moving beyond simple human alignment to create targeted challenge sets that can expose undue preferences.

\section*{Limitations}

Our study is characterized by several methodological constraints that merit critical reflection. We limited our investigation to sentence-level MT, which, while representative of prevalent machine translation scenarios, does not cover phenomena potentially arising within more extended contexts, e.g., document-level \cite{jiang-etal-2022-blonde, raunak-etal-2023-evaluating,fernandes2025llms} and conversational MT \cite{agrawal_2024_context_helpful_chat_translation}. The English source sentences examined were derived from naturalistic passages of high quality and relatively limited syntactic variation. 

Moreover, our evaluation of neural QE metrics with extra-sentential context, relies on inference-time strategies for integrating contextual information~\cite{vernikos-etal-2022-embarrassingly,agrawal_2024_context_helpful_chat_translation}. We acknowledge that translating context may introduce additional bias depending on the specific translation system used. 
We leave a more in-depth investigation of this phenomenon---including more resource-intensive solutions, such as targeted fine-tuning or interpretability-driven strategies for selecting few-shot examples to mitigate biases (e.g., gender) in the translation of contextual information~\cite{thakur-etal-2023-language-gender-makeover-few-shot-interventions,zaranis-etal-2024-analyzing}---to future work.

Finally, methodologically, we implemented GEMBA-DA in a zero-shot configuration, aligning with contemporary research interested in pushing zero-shot problem-solving \citep{guo2025deepseek}. 
We defer comprehensive exploration of few-shot and MQM-based prompting strategies \citep{kocmi-federmann-2023-gemba} to future work.

\section*{Ethical Considerations}

Using gender as a variable requires careful ethical considerations.
The corpora we used, initially annotated for gender identity using textual markers like pronouns and titles, present two main concerns.
First, gender is a nuanced, evolving attribute, making reliance on textual information alone potentially inaccurate, especially with unverifiable or outdated sources.
Second, much of our analysis uses binary gender categories (masculine and feminine). However, we recognize gender as a spectrum and advocate for future studies to account for non-binary and other gender identities. To support gender inclusivity in machine translation, we assessed quality estimation on gender-neutral translations in Italian, Spanish, and German using the mGeNTE corpus.

\section*{Acknowledgements}
We would like to thank Ben Peters, Dennis Fucci, António Farinhas, Sonal Sannigrahi for their feedback and insightful discussions in the earlier versions of the paper, and Anna Currey for providing valuable help with the MT-GenEval dataset.
This work was supported by the Portuguese Recovery and Resilience Plan through project C645008882-00000055 (Center for Responsible AI), by EU's Horizon Europe Research and Innovation Actions (UTTER, contract 101070631), by the project DECOLLAGE (ERC-2022-CoG 101088763), and by FCT/MECI through national funds and when applicable co-funded EU funds under UID/50008: Instituto de Telecomunicações.

\bibliography{custom}

\clearpage
% \onecolumn
\appendix
\section{Experimental Details}
\label{sec:app:experimental_details}

\subsection{Computing DA Scores}
\label{ssec:app:computing_DA_scores}

\paragraph{COMET.} We use the HuggingFace Hub for all COMET metrics and the official implementation from \url{https://github.com/Unbabel/COMET}.
All metrics produce an inclusive score between 0 and 1, indicating the worst and best translation possible.

\paragraph{MetricX 23.} We use the official implementation found at \url{https://github.com/google-research/metricx}. Since these metrics produce error assessments, with a score between 0 (no errors) and 25 (maximum errors), we rescale the output as $QE(\cdot) = 1 - o(\cdot) /\ 25$, where $o(\cdot)$ is the raw output of the metric.

\paragraph{GEMBA.} We use the same decoding configuration and post-processing for all the language models prompted with GEMBA-like prompts. Specifically, we use greedy decoding, with a maximum number of generated tokens of 256. Each model's specific chat template is retrieved from the transformers' implementation \citep{wolf-etal-2020-transformers}.  We use the instruction-tuned variants and a zero-shot prompt for all open-weight models. Non-contextual and context-aware prompts are reported in Table~\ref{tab:gemba_prompts}.
We experimented two variants to include context in prompts, finding no significant difference among those. We finally used the v1 reported in Table~\ref{tab:gemba_prompts}.

We post-process each output using regexes to capture the recommended score, e.g., \texttt{``i would score this translation (\textbackslash d+(\textbackslash.\textbackslash d+)?).''} Please refer to our repository for the full list of pattern strings. If no patterns are matched, we take the first integer number in the response. Since some models produced numbered lists (e.g., ``1. [...]''), we filter out every row with a numerical score lower than ten or without a score.
As prompts request a score between 0 and 100, we rescale them to the interval 0 to 1 after extraction. Using these filters, we drop less than 1\% of the rows on average.  

\noindent
The Hugging Face checkpoint and OpenAI's model IDs are reported in Table~\ref{tab:hf_ids}.

\begin{table*}[!t]
\small 
\centering
\setlength\tabcolsep{2pt}
% \resizebox{\linewidth}{!}{%
\begin{tabular}{lll}
\toprule
\textbf{\texttt{[1]}} & CometKiwi~22 & \texttt{Unbabel/wmt22-cometkiwi-da} \\ 
\textbf{\texttt{[2]}} & CometKiwi~23~XL & \texttt{Unbabel/wmt23-cometkiwi-da-xl} \\ 
\textbf{\texttt{[3]}} & CometKiwi~23~XXL & \texttt{Unbabel/wmt23-cometkiwi-da-xxl} \\ 
\textbf{\texttt{[4]}} & xCOMET~XL & \texttt{Unbabel/XCOMET-XL} \\ 
\textbf{\texttt{[5]}} & xCOMET~XXL & \texttt{Unbabel/XCOMET-XXL} \\ 
\textbf{\texttt{[6]}} & MetricX~23~L & \texttt{google/metricx-23-qe-large-v2p0} \\ 
\textbf{\texttt{[7]}} & MetricX~23~XL & \texttt{google/metricx-23-qe-xl-v2p0} \\ 
\midrule
\textbf{\texttt{[8]}} & Mistral~7B & \texttt{mistralai/Mistral-7B-Instruct-v0.2} \\ 
\textbf{\texttt{[9]}} & Gemma~2~9B & \texttt{google/gemma-2-9b-it} \\ 
\textbf{\texttt{[10]}} & Llama~3.1~70B & \texttt{meta-llama/Meta-Llama-3.1-70B-Instruct} \\ 
\textbf{\texttt{[11]}} & GPT 4o & \texttt{gpt-4o-2024-05-13} \\ 
\bottomrule
\end{tabular}  
% }
\caption{\textbf{Hugging Face Hub ID} for each QE metrics tested.}
\label{tab:hf_ids}
\end{table*}

\begin{table*}[!t]
\centering
\small
% \scalebox{0.9}{
\begin{tabular}{@{}p{3cm}p{12cm}lcc@{}}
\toprule
\textbf{Format} & \textbf{Template} \\ \midrule
GEMBA & \texttt{Score the following translation from \{src\_lang\} to \{tgt\_lang\} on a continuous scale from 0 to 100, where score of zero means ``no meaning preserved'' and score of one hundred means ``perfect meaning and grammar''.\newline
\{src\_lang\} source: ``\{source\}''\newline
\{tgt\_lang\} translation: ``\{hyphothesis\}''\newline
Score: }\\ \midrule
GEMBA\textsubscript{ctx} v1 & \texttt{Score the following translation from \{src\_lang\} to \{tgt\_lang\} on a continuous scale from 0 to 100 \textbf{given the preceding context}, where score of zero means ``no meaning preserved'' and score of one hundred means ``perfect meaning and grammar''.\newline
\{src\_lang\} source: ``\{context\}''\newline
\{src\_lang\} source: ``\{source\}''\newline
\{tgt\_lang\} translation: ``\{hyphothesis\}''\newline
Score: }\\
GEMBA\textsubscript{ctx} v2 & \texttt{Score the following translation from \{src\_lang\} to \{tgt\_lang\} on a continuous scale from 0 to 100, where score of zero means ``no meaning preserved'' and score of one hundred means ``perfect meaning and grammar''.\newline
\textbf{You can use the preceding context to evaluate the translation of the source.}
\{src\_lang\} source: ``\{context\}''\newline
\{src\_lang\} preceding context: ``\{source\}''\newline
\{tgt\_lang\} translation: ``\{hyphothesis\}''\newline
Score: }\\ \bottomrule
% HellaSwag &  &  &  \\
\end{tabular}
% }
\caption{\textbf{Templates for LMs.} We use GEMBA-like \citep{kocmi-federmann-2023-large} prompts to evaluate context-free (top) and context-aware (bottom) translation quality. Edits to the original prompt to account for context are in bold. We used v1 for all experiments.}
\label{tab:gemba_prompts}
\end{table*}

\subsection{Statistical Significance}
\label{ssec:statistical_significance}
We confirm the validity of our results using statistical significance tests. Particularly 
for testing the statistical significance of the error rate ratio between the gender groups ($\Phi(S^F,S^M) \rightarrow1$), we use paired bootstrap resampling ($p<.05$). Similarly for the statistical significance of QE ratios, we use the one-sample t-test ($p<.05$).

\subsection{Filtering Details for Machine Translation Quality Experiments}\label{app:sec:GT_filtering_details}

As outlined in~\ref{subsec:findings_GT_translations}, to explore whether QE metrics exhibit gender bias when assessing machine-generated translations, we translate the source-feminine ($s_F$) and source-masculine ($s_M$) texts from the counterfactual subset of MTGenEval with Google Translate (GT). Then the counterfactual pairs, ($s_F, \mathrm{GT_F}$) and ($s_M, \mathrm{GT_M}$) are filtered through the following two-step process:
\begin{enumerate}
    \item \textbf{Target gender inflection:} We exclude counterfactual pairs where either of the translations do not exhibit the correct target gender inflection, as categorized by \citet{currey-etal-2022-mt}.
    \item \textbf{Translation quality consistency:} We remove pairs with translations of different quality, as measured by BLEU \cite{papineni-etal-2002-bleu}. Specifically, we split translations into 5 quality categories (``Poor'',``Fair'',``Good'',``Very Good'' and ``Excellent''), following~\citet{lavie-2011-evaluating}. We retain only the pairs of instances for which both translations belong to the same quality category. Additionally, within each quality category, we confirm that BLEU does not significantly differ between gender groups using the Wilcoxon test ($p < .05$).
\end{enumerate}
This filtering process ensures that all retained counterfactual pairs are correctly gender-inflected and of equivalent translation quality. Importantly, we retain around half of the initial instances as indicated in Table~\ref{tab:filter_stats}, most of which are of high-quality according to BLEU.

\subsection{Hardware specifications}
All our experiments were conducted using 2 NVIDIA RTX A6000 GPUs.

\subsection{Discussion on artifacts}
In our analysis we use the  MT-GenEval \citep{currey-etal-2022-mt}, GATE \citep{Rarrick2023GATEAC}, and mGeNTE \citep{savoldi2025mgentemultilingualresourcegenderneutral} datasets  and they can be freely used for research purposes as they are under the CC-BY-SA-3.0, MIT, and 
 CC-BY-4.0 licenses respectively.

\section{Additional Results}\label{app:sec:additional_results}

This section reports additional figures and tables to supplement the findings in the main body.

\subsection{Results on Gender Ambiguous Instances}
\label{app:ssec:results_gender_ambiguous}
\begin{figure*}[!th]
    \centering 
    \includegraphics[width=0.84\linewidth]{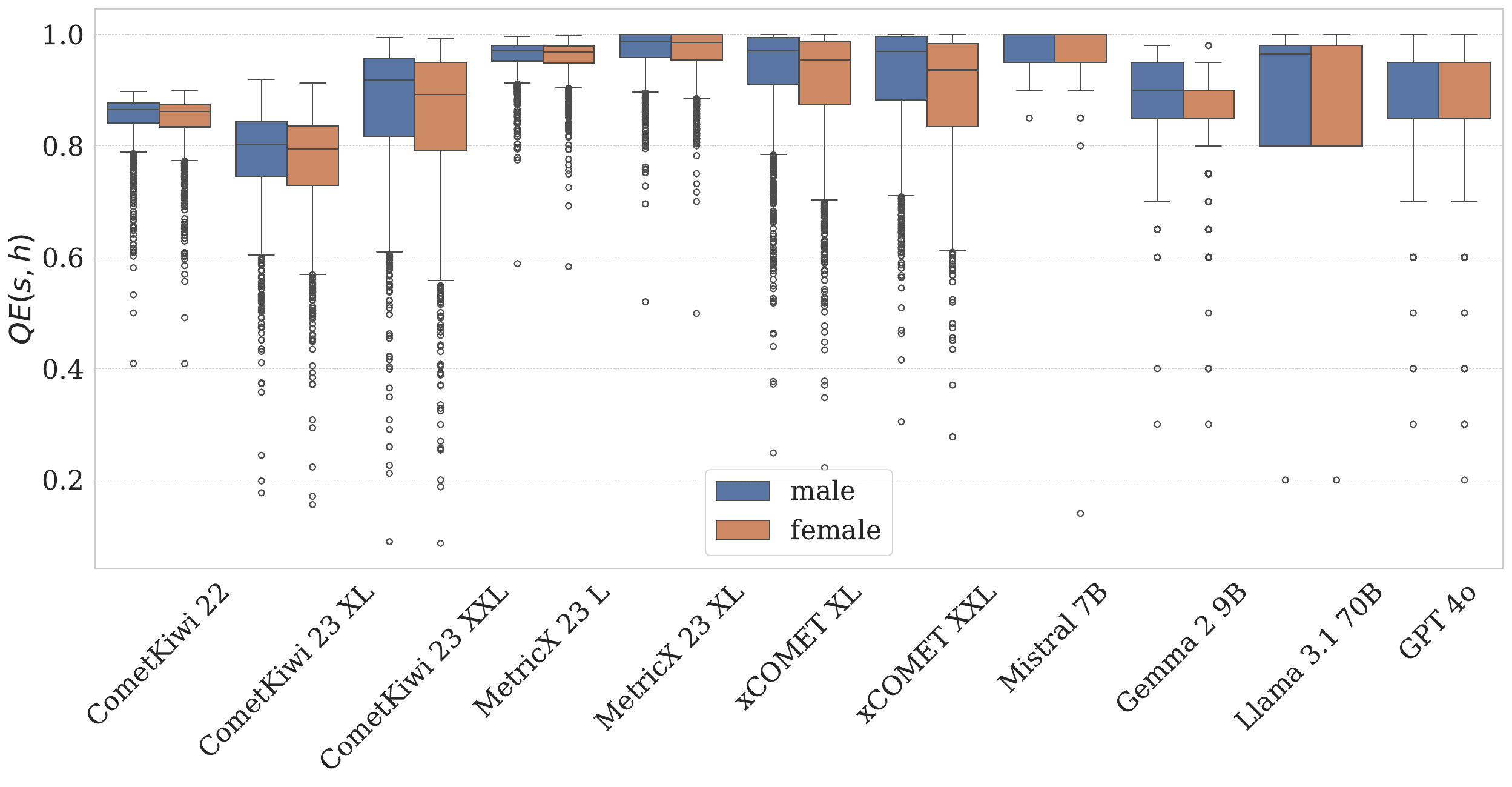}
    % \vspace{-10pt}
    \caption{\textbf{Raw QE scores for all QE metrics on MT-GenEval.} Ambiguous case. Contextual subset, test instances. Averaged results across languages.}\label{fig:raw_scores_MTGenEval_contextual_ambiguous}
\end{figure*}

\begin{figure*}[!th]
    \centering
    % \begin{subfigure}[b]{0.90\linewidth}
        % \centering
    \includegraphics[width=0.84\linewidth]{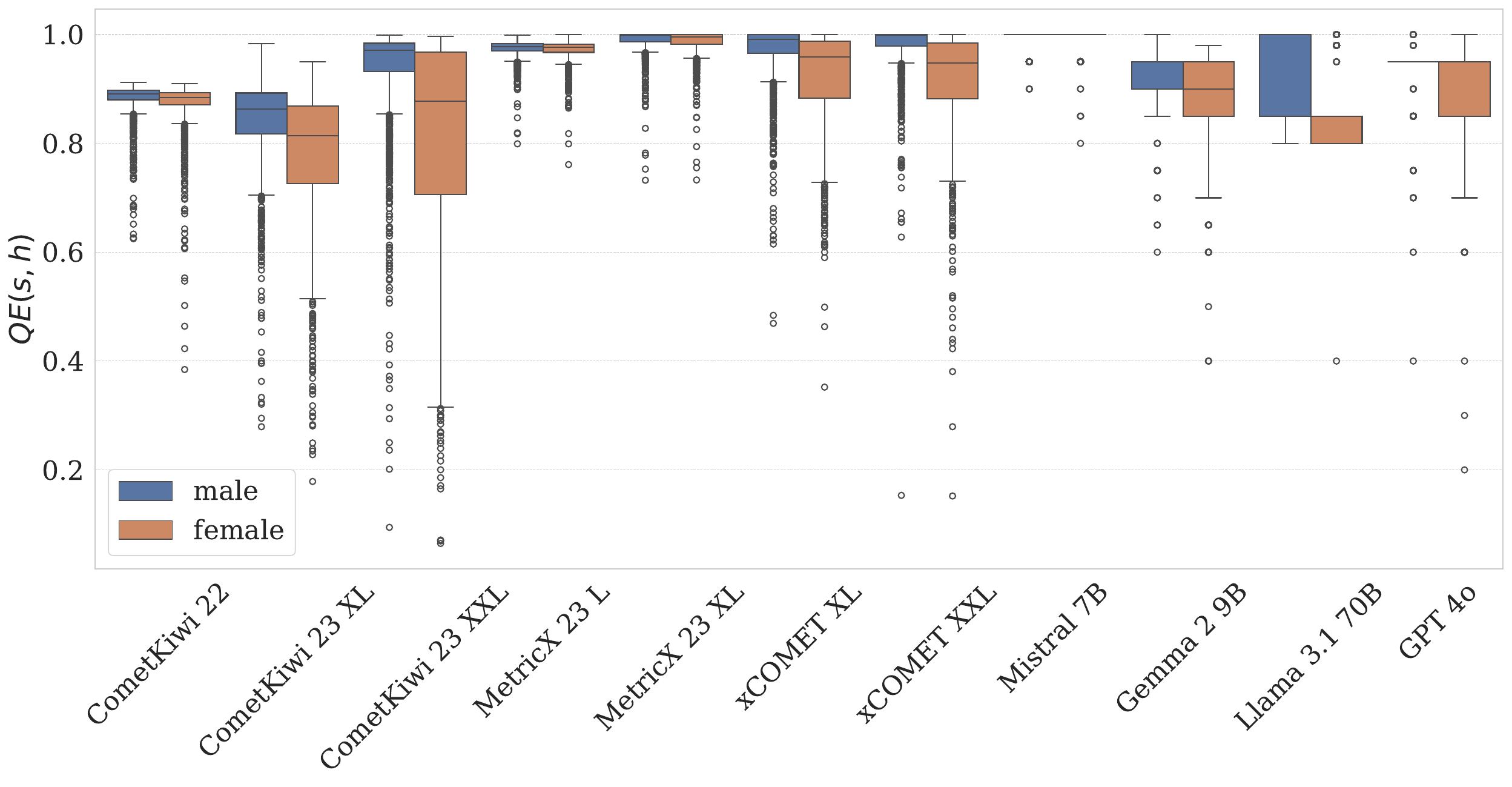}
    % \end{subfigure}

    \caption{\textbf{Raw QE scores for all QE metrics on GATE.} \textit{Ambiguous case.} Averaged results across three languages (fr, es, and it).}
    \label{fig:raw_scores_GATE}
\end{figure*}

\begin{figure*}[!th]
    \centering
    % \begin{subfigure}{0.92\linewidth}
        % \centering
    \includegraphics[width=0.84\linewidth]{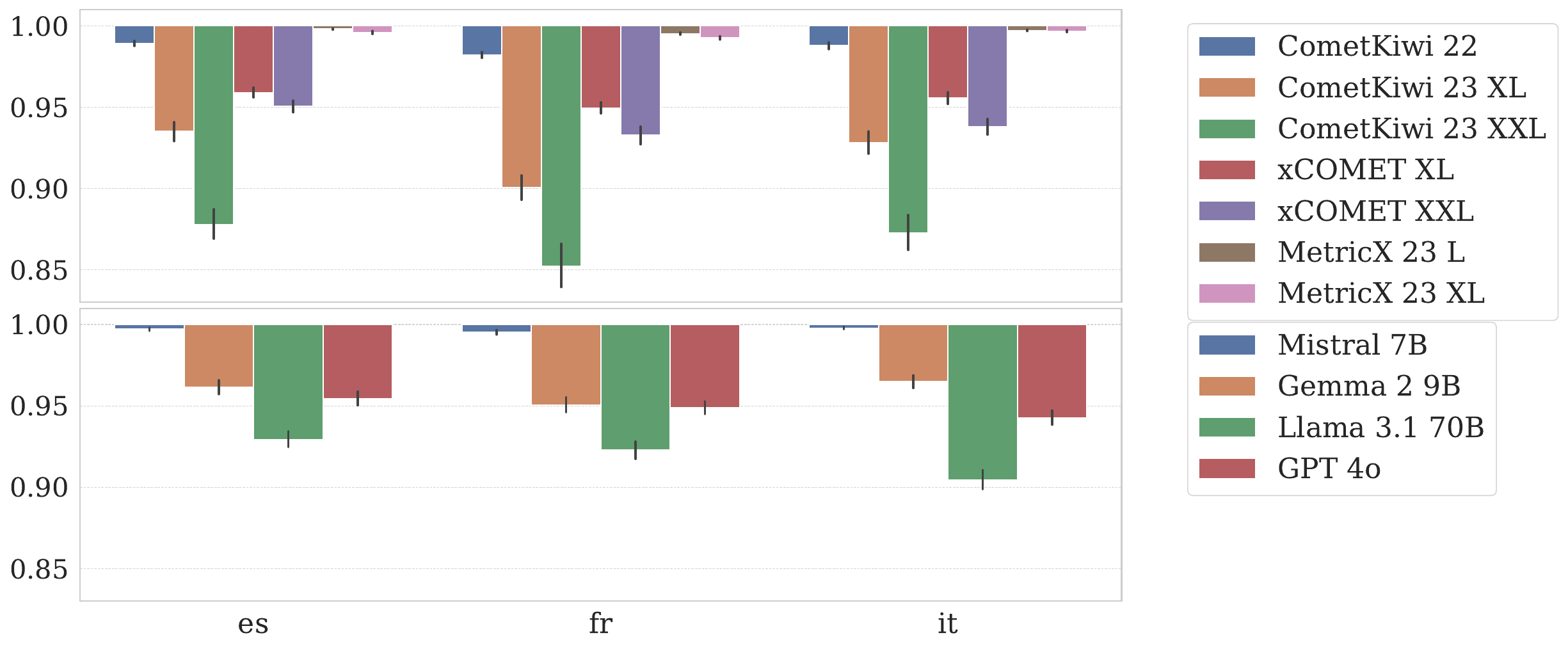}
    % \end{subfigure}
    \caption{\textbf{Ratio $QE(s,h_F) /\ QE(s,h_M)$ on GATE.} \textit{Ambiguous case.} Average and 95\% confidence interval on the test set by language. Neural QE metrics (top) and GEMBA-prompted language models (bottom). 
    }
    \label{fig:ratio_GATE}
\end{figure*}

\begin{figure*}[!th]
    \centering
    \includegraphics[width=\linewidth]{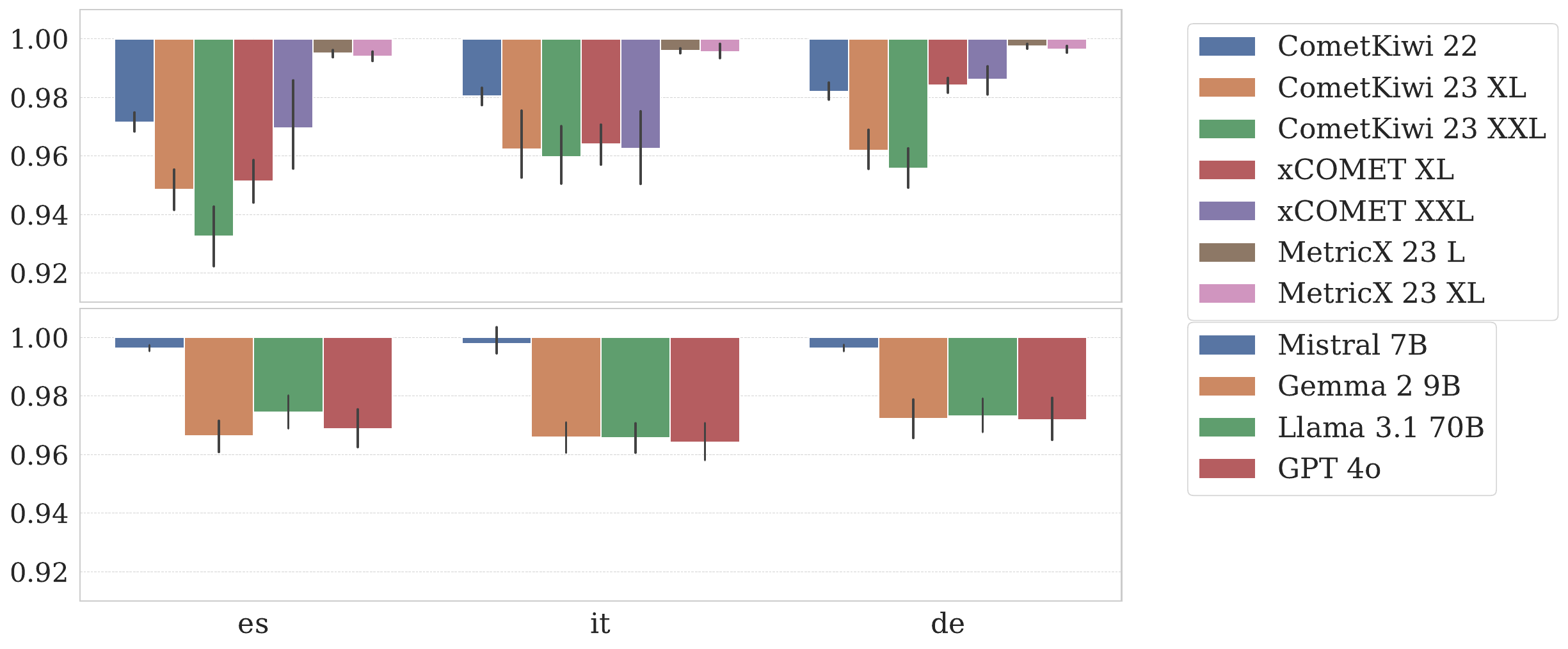}
    \caption{\textbf{Ratio $\mathrm{QE}(s,h_N) /\ \mathrm{QE}(s,h_G)$ on mGeNTE (Set-N).} \textit{Ambiguous case.} $h_N$ and $h_G$ denote the neutral and gendered translations, respectively. Average and 95\% confidence interval on the test set.}
    \label{fig:ratio_mGeNTE}
\end{figure*}

In this part, we provide a comprehensive view of results for the ambiguous case. Specifically, Figures~\ref{fig:raw_scores_MTGenEval_contextual_ambiguous} and~\ref{fig:raw_scores_GATE} report the raw QE scores when assessing masculine and feminine references, on MT-GenEval and GATE respectively. In addition, Figures~\ref{fig:ratio_GATE} and~\ref{fig:ratio_mGeNTE} report the ratios of the QE scores on GATE and mGeNTE datasets respectively.

\subsection{Results on Gender Unambiguous Instances}
\label{app:ssec:results_gender_unambiguous}

\paragraph{Additional results for Intra-sentential Case.}
In this setup, \textit{gender ambiguity is resolved through contextual cues within the sentence itself}. To provide a more comprehensive analysis, we include additional results for this case. Table~\ref{app:tab:results_counterfactual_ref_full_table} presents the complete results, including the total error rate $\mathrm{ER}$, error rate ratio between gender groups $\Phi(S^F,S^M)$, and QE score ratios $\mathrm{QE}(s_F,h_F) /\ \mathrm{QE}(s_M,h_M)$, averaged across languages. Additionally, Tables~\ref{app:tab:counterfactual_stat_tests_ratio_Phi} and~\ref{app:tab:counterfactual_stat_tests_ratio_QE} provide the per-language error rate ratios between gender groups $\Phi(S^F,S^M)$ and  per-language QE score ratios $\mathrm{QE}(s_F,h_F) /\ \mathrm{QE}(s_M,h_M)$, along with the corresponding results of statistical significance tests.

\paragraph{Additional results for Extra-sentential Case.}
In this setup, \textit{gender ambiguity is resolved by contextual cues from preceding sentences}.  As explained in \S\ref{subsec:formalizing_qe}, we experiment with two variants for neural QE metrics: 1) prepending the context of the source \textit{unmodified} to the candidate hypothesis, and 2) prepending the  context \textit{translated} to the candidate hypothesis. For GEMBA metrics we experiment with two versions for including context(see Table~\ref{tab:gemba_prompts}). 

To provide a comprehensive view, we include the complete results. Table~\ref{app:tab:results_contextual_non_amb_aggr} presents the complete results, including the total error rate $\mathrm{ER}$, the error rates for sources with masculine referents $\mathrm{ER}(S^M)$ and feminine referents $\mathrm{ER}(S^F)$, and the error rate ratio between gender groups $\Phi(S^F,S^M)$, averaged across languages. Table~\ref{app:tab:contextual_non_amb_stat_tests_phi} provides the  error rate ratios between gender groups $\Phi(S^F,S^M)$, along with the corresponding results of statistical significance tests for individual languages. In addition, Table~\ref{app:tab:ties_contextual_non_amb_llms} summarizes the aggregated tie rates across all languages. 

According to our findings, the majority of QE metrics is biased. Additionally, translating the context versus prepending it unmodified to the hypothesis, might improve accuracy but in the trade of bias.

\begin{figure*}[h]
    \centering
    \includegraphics[width=0.6\linewidth]{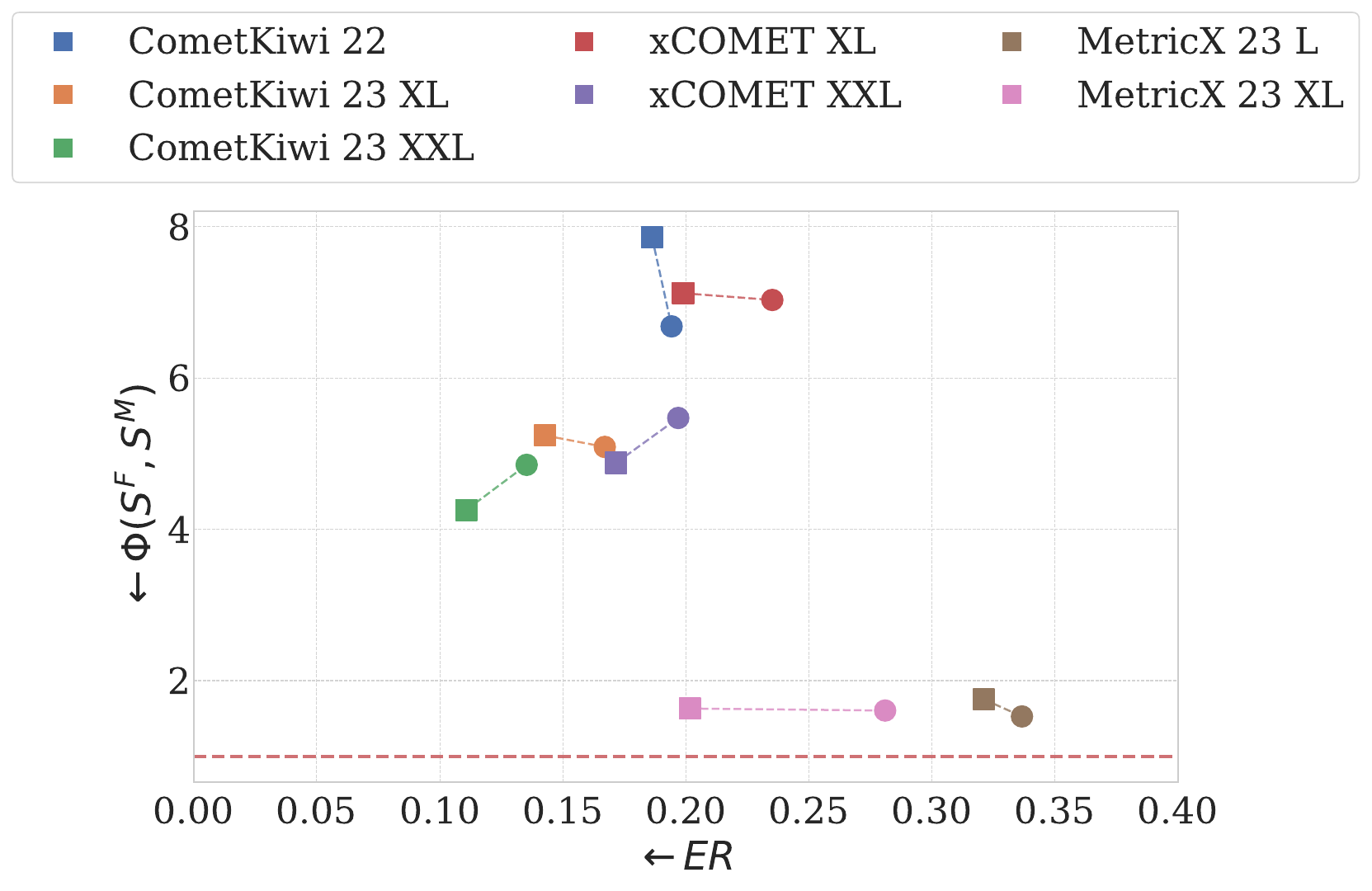}
    \caption{\textbf{Total error rate $ER$ and error rate ratio $\Phi(S^F,S^M)$ between groups for neural QE variants that use translated context.} The translated context is generated by GT (\ding{110}) or NLLB (\ding{108}). Unambiguous, Extra-sentential case. Instances of MTGenEval's counterfactual subset .  Mean values across eight languages. The red dotted horizontal line indicates parity for $\Phi$.}
    \label{fig:comparison_context_trans_quality}
\end{figure*}

Lastly, as indicated in  Figure~\ref{fig:comparison_context_trans_quality}, the translation system being used to translate the context---Google Translate (\ding{110}) or NLLB (\ding{108})---does not affect any trend majorly. Metrics might show marginal improvements in accuracy or gender equality, however the majority of them is biased.

\subsection{Results when assessing QE metrics on Automatic Translations}
In \S~\ref{subsec:findings_GT_translations} we explore whether QE metrics demonstrate any systematic biases when applied to automatic translations. 
As detailed in \S~\ref{subsec:findings_GT_translations} 
we focus on the intra-sentential case, where the gender identity of the referent is resolved by the source sentence itself. We conduct our analysis on counterfactual pairs ($s_F, \mathrm{GT_F}$) and ($s_M, \mathrm{GT_M}$), by combining the counterfactual subset of MTGenEval and outputs obtained from Google Translate (see Appendix~\ref{app:sec:GT_filtering_details}). 

To provide a comprehensive view, we include the complete results on individual language pairs. Specifically, 
 in Table~\ref{app:tab:counterfactual_stat_tests_ratio_QE_GT}, we report the QE score ratios, along with the corresponding statistical tests, for individual languages. In addition, Table~\ref{app:tab:counterfactual_stat_tests_ratio_Phi_GT} presents the per-language error rate ratios $\Phi_{GT}(S^F,S^M)$ between gender groups. As shown, when QE metrics are challenged to select the correct gender inflection for an automatic translation, they demonstrate significantly higher error rates on sources with feminine referents than on their masculine counterparts. A finding inline with our observations when assessing human-written translations.

\section{AI Assistants}
We have used Github Copilot\footnote{https://github.com/features/copilot} and Claude 3.5 Sonnet during development of our research work.

\begin{table*}[!th]
\small
\centering
\begin{tabular}{lccc}
\toprule
\textbf{Metric} & \textbf{$\mathrm{ER}$} & \textbf{$\Phi(S^F,S^M)$}  &  $\mathrm{QE}(s_F,h_F) /\ \mathrm{QE}(s_M,h_M)$ \\
\midrule
CometKiwi 22  & 0.11 & 1.70 & 0.997\\
CometKiwi 23 XL & 0.09 & 1.18 & 1.001 \\
CometKiwi 23 XXL & \textbf{0.07} & \textbf{0.87} & 1.003\\
xCOMET XL & 0.10 &  1.81 & 0.988\\
xCOMET XXL & 0.08& 1.32 & 0.995\\
MetricX 23 L & 0.31 & 1.25 & 0.998\\
MetricX 23 XL  & 0.12 & 1.19& \textbf{1.000}\\
\midrule
Llama 3.1 70B & 0.31 & 1.16 & 0.995 \\
Gemma 2 9B & 0.28 & 1.36 &  \textbf{1.000}\\
Mistral 7B & 0.74 & 1.13 &  \textbf{1.000}\\
GPT 4o & 0.16 & 1.15 & 1.001 \\
\bottomrule
\end{tabular}
\caption{\textbf{Total error rate ($\mathrm{ER} \downarrow$), error rate ratio between gender groups ($\Phi(S^F,S^M) \rightarrow1$) and QE score ratios}. Unambiguous, Intra-sentential case. Counterfactual subset of MT-GenEval. Mean results across eight languages.}
\label{app:tab:results_counterfactual_ref_full_table}
\end{table*}

\begin{table*}[!th]
\centering
\small
\begin{tabular}{lcccccccc}
\toprule
\rowcolor{white}
\textbf{Metrics} & \multicolumn{8}{c}{$\Phi(S^F,S^M) = ER(S^F) /\ ER(S^M)$} \\
& de & es & fr & it & pt & ru & hi & ar \\
\midrule
CometKiwi 22 & \significant 2.500 & \significant 1.909 & \significant 2.333 & \significant 1.773 & 0.938  & \significant 2.500 & 0.683  & 1.000 \\
CometKiwi 23 XL & \significant 1.143 & \significant 1.136 & 0.500  & 1.000  & \significant 1.333 & \significant 2.333 & \significant 1.297 & 0.706  \\
CometKiwi 23 XXL & 0.286  & 0.923  & 0.750  & \significant 1.036 & 0.692  & \significant 2.000 & 1.000  & 0.273  \\
\addlinespace[0.2cm]

xCOMET XL & \significant 3.000 & \significant 1.154 & \significant 2.200 & \significant 1.346 & \significant 2.429 & \significant 2.200 & \significant 1.181 & 1.000  \\
xCOMET XXL & 0.857  & \significant 1.333 & \significant 1.143 & \significant 1.321 & \significant 1.111  & \significant 2.500 & \significant 1.426 & 0.833  \\
\addlinespace[0.2cm]

MetricX 23 L & \significant 1.037 & \significant 1.582 & \significant 1.459 & \significant 1.186 & 0.839  & \significant 1.302 & 0.937  & \significant 1.660 \\
MetricX 23 XL & \significant 1.600 & \significant 1.083 & 0.857  & 0.959  & 0.550  & \significant 2.182 & \significant 1.182 & \significant 1.130 \\
\addlinespace[0.2cm]

Mistral 7B & \significant 1.190 & \significant 1.125 & \significant 1.184 & \significant 1.194 & \significant 1.098 & \significant 1.191 & \significant 1.045 & 0.990  \\
Gemma 2 9B & 0.775  & \significant 1.317 & \significant 1.767 & \significant 1.443 & \significant 1.459 & \significant 1.421 & \significant 1.248 & \significant 1.426 \\
Llama 3.1 70B & 0.902  & \significant 1.265 & \significant 1.113 & \significant 1.161 & \significant 1.140 & \significant 1.185& \significant 1.304 & \significant 1.210 \\
GPT 4o & 0.857  & \significant 1.200 & 0.939  & \significant 1.167 & \significant 1.037 & \significant 1.368 & \significant 1.214 & \significant 1.400 \\

\bottomrule
\end{tabular}
\caption{ \textbf{Per language error rate ratio $\Phi(S^F,S^M)$ between gender groups} along with the corresponding statistical significance tests. \colorbox{red!20}{Highlighted values} indicate that error rates for sources with feminine referents  are significantly higher than for their masculine counterparts. Unambiguous, Intra-sentential case. Counterfactual subset of MT-GenEval.}
\label{app:tab:counterfactual_stat_tests_ratio_Phi}
\end{table*}

\begin{table*}
\centering
\small
\begin{tabular}{lcccccccc}
\toprule
\rowcolor{white}
\textbf{Metrics} & \multicolumn{8}{c}{$QE(s_F,h_F) / QE(s_M,h_M)$ } \\
& de & es & fr & it & pt & ru & hi & ar \\
\midrule

CometKiwi 22 & \significant 0.995 & \significant 0.995 & \significant 0.995 & \significant 0.996 & 1.000  & \significant 0.997 & \significant 0.998 & 1.000  \\
CometKiwi 23 XL & 0.997  & 1.000  & 0.998  & 0.996  & 0.998  & 1.000  & 1.006  & 1.015  \\
CometKiwi 23 XXL & 1.005  & 0.995  & 1.006  & \significant 0.991 & 0.999  & 1.001  & 1.006  & 1.025  \\
\addlinespace[0.2cm]

xCOMET XL & \significant 0.988 & \significant 0.975 & 0.997  & \significant 0.979 & \significant 0.982 & \significant 0.991 & 0.996  & 0.995  \\
xCOMET XXL & \significant 0.997 & 0.993  & 0.997  & \significant 0.989 & 1.001  & 0.998  & \significant 0.989 & 0.995  \\
\addlinespace[0.2cm]

MetricX 23 L & 0.999  & 0.998  & \significant 0.998 & \significant 0.998 & 1.000  & 0.999  & 1.000  & \significant 0.992 \\
MetricX 23 XL & \significant 0.998 & 0.998  & 0.998  & 1.000  & 1.001  & 1.000 & 1.002  & 1.000 \\
\addlinespace[0.2cm]

Llama 3.1 70B & 1.002  & 0.993  & \significant 0.989 & 0.999  & 0.993  & \significant 0.991 & 0.995  & 1.001  \\
Gemma 2 9B & 0.997  & 1.001  & 1.002  & 1.004  & 0.999  & 0.998  & 1.003  & 0.999  \\
\addlinespace[0.2cm]

Mistral 7B & 1.000  & 1.000  & 0.999  & 1.000  & \significant 0.998 & 1.000  & 1.002  & 1.002  \\
GPT 4o & 0.999  & 1.003  & 0.998  & 1.003  & 1.000  & 0.993  & 1.000  & 1.007  \\

\bottomrule
\end{tabular}
    \caption{\textbf{Per language QE score ratios} along with the corresponding statistical significance tests when assessing correctly-inflected feminine and masculine references. \colorbox{red!20}{Highlighted values} indicate that correctly-inflected feminine references receive significantly lower scores compared to their masculine counterparts.  Unambiguous, Intra-sentential case. Counterfactual subset of MT-GenEval.}
    \label{app:tab:counterfactual_stat_tests_ratio_QE}
\end{table*}

\begin{table*}[h]
    \centering
    \resizebox{\textwidth}{!}{%
    
    \begin{tabular}{l:cccccc:cccc:cccc:ccccccc}
    \rowcolor{black!2}&&&&&&&&&&&&&&&&&&&&&\\
    \rowcolor{black!2}
    Metric & \rotatebox{90}{CometKiwi 22 } & \rotatebox{90}{CometKiwi 22 } & \rotatebox{90}{CometKiwi 23 XL } & \rotatebox{90}{CometKiwi 23 XL } & \rotatebox{90}{CometKiwi 23 XXL } & \rotatebox{90}{CometKiwi 23 XXL } & \rotatebox{90}{xCOMET XL } & \rotatebox{90}{xCOMET XL } & \rotatebox{90}{xCOMET XXL } & \rotatebox{90}{xCOMET XXL } & \rotatebox{90}{MetricX 23 L } & \rotatebox{90}{MetricX 23 L } & \rotatebox{90}{MetricX 23 XL } & \rotatebox{90}{MetricX 23 XL } & \rotatebox{90}{Llama 3.1 70B } & \rotatebox{90}{Llama 3.1 70B V2} & \rotatebox{90}{Gemma 2 9B } & \rotatebox{90}{Gemma 2 9B V2} & \rotatebox{90}{Mistral 7B } & \rotatebox{90}{Mistral 7B V2} & \rotatebox{90}{GPT- 4o} \\ 
    \hdashline
    \rowcolor{black!2}
        $\mathrm{tr}(c)=\checkmark$ & \rotatebox{90}{} & \rotatebox{0}{\checkmark} & \rotatebox{90}{} & \rotatebox{0}{\checkmark} & \rotatebox{90}{} & \rotatebox{0}{\checkmark} & \rotatebox{90}{} & \rotatebox{0}{\checkmark} & \rotatebox{90}{} & \rotatebox{0}{\checkmark} & \rotatebox{90}{} & \rotatebox{0}{\checkmark} & \rotatebox{90}{} & \rotatebox{0}{\checkmark} & \rotatebox{0}{} & \rotatebox{0}{} & \rotatebox{0}{} & \rotatebox{0}{} & \rotatebox{0}{} & \rotatebox{0}{}  & \rotatebox{0}{} \\ 
    \hdashline
    $\mathrm{ER}$ & 0.28 & 0.19 & 0.28 & 0.17 & \textbf{0.14} & \textbf{0.14} & 0.29 & 0.24 & 0.20 & 0.20 & 0.39 & 0.34 & 0.22 & 0.28 & 0.61 & 0.60 & 0.57 & 0.59 & 0.88 & 0.86 & 0.44 \\
$ER(S^M)$ & 0.32 & 0.10 & 0.30 & 0.09 & 0.09 & 0.08 & 0.11 & 0.09 & \textbf{0.07} & 0.09 & 0.46 & 0.28 & 0.17 & 0.24 & 0.38 & 0.38 & 0.31 & 0.34 & 0.79 & 0.75 & 0.24 \\
$ER(S^F)$ & 0.25 & 0.28 & 0.26 & 0.24 & \textbf{0.19} & \textbf{0.19} & 0.45 & 0.37 & 0.32 & 0.30 & 0.33 & 0.39 & 0.26 & 0.32 & 0.81 & 0.81 & 0.81 & 0.82 & 0.96 & 0.95 & 0.61 \\
$\Phi(S^F,S^M)$& 0.84 & 6.68 & \textbf{0.92} & 5.09 & 2.20 & 4.85 & 4.53 & 7.03 & 4.97 & 5.47 & 0.73 & 1.53 & 1.54 & 1.60 & 2.20 & 2.22 & 2.71 & 2.54 & 1.22 & 1.29 & 2.64 \\
    \bottomrule
    \end{tabular}%
    }
\caption{\textbf{Total error rate $\mathrm{ER}$, error rate per-group $\mathrm{ER}(S^M)$ , $\mathrm{ER}(S^F)$ and error rate ratio between gender groups $\Phi(S^F,S^M)$} for all metric versions examined.  Unambiguous, Extra-sentential case.
Contextual subset of MT-GenEval. $\mathrm{tr}(c)=\checkmark$: context-aware metrics with translated context. Results are averaged across all languages. }
\label{app:tab:results_contextual_non_amb_aggr}
\end{table*}

\begin{table*}
\centering
\small
% \scalebox{0.8}{
\begin{tabular}{lccccccccc}
\toprule
\rowcolor{white}
\textbf{Metrics} &  $\mathrm{tr}(c)$ &  \multicolumn{8}{c}{$\Phi(S^F,S^M) = ER(S^F) / ER(S^M)$ } \\
&& de & es & fr & it & pt & ru & hi & ar \\
\midrule

CometKiwi 22  &  & 0.691  & 0.853  & 0.895  & \significant 1.117 & \significant 1.068 & 0.724  & 0.638  & 0.704  \\
CometKiwi 22  &\checkmark & \significant 6.636 & \significant 9.068 & \significant 8.990 & \significant 8.278 & \significant 7.364 & 1\significant 0.820 & \significant 1.506 & 0.807  \\
\addlinespace[0.2cm]

CometKiwi 23 XL  &  & \significant 1.066 & \significant 1.012 & \significant 1.133 & \significant 1.049 & \significant 1.248 & 0.744  & 0.490  & 0.595  \\
CometKiwi 23 XL  &\checkmark & \significant 4.771 & \significant 4.610 & \significant 4.362 & \significant 5.371 & \significant 6.102 & 1\significant 0.396 & \significant 4.371 & 0.736  \\
\addlinespace[0.2cm]

CometKiwi 23 XXL  & & \significant 2.675 & \significant 2.597 & \significant 2.221 & \significant 2.161 & \significant 2.627 & \significant 2.027 & \significant 2.092 & \significant 1.162 \\
CometKiwi 23 XXL  &\checkmark & \significant 3.487 & \significant 3.656 & \significant 3.508 & \significant 6.821 & \significant 6.496 & \significant 9.941 & \significant 4.168 & 0.746  \\
\addlinespace[0.2cm]

xCOMET XL  & & \significant 5.887 & \significant 4.686 & \significant 4.161 & \significant 4.859 & \significant 4.308 & \significant 6.104 & \significant 4.611 & \significant 1.622 \\
xCOMET XL  &\checkmark & 1\significant 0.291 & \significant 4.668 & 1\significant 0.043 & \significant 7.486 & \significant 8.606 & \significant 7.320 & \significant 6.547 & \significant 1.300 \\
\addlinespace[0.2cm]

xCOMET XXL  &  & \significant 4.461 & \significant 6.011 & \significant 5.759 & \significant 6.337 & \significant 4.848 & \significant 3.921 & \significant 5.675 & \significant 2.760 \\
xCOMET XXL  &\checkmark  & \significant 3.849 & \significant 5.810 & \significant 8.804 & \significant 6.020 & \significant 5.934 & \significant 6.153 & \significant 6.265 & 0.945  \\
\addlinespace[0.2cm]

MetricX 23 L  & & 0.559  & 0.681  & 0.650  & 0.821  & 0.837  & 0.864  & 0.823  & 0.607  \\
MetricX 23 L  &\checkmark & \significant 1.118 & \significant 1.476 & \significant 1.649 & \significant 1.826 & \significant 1.050 & \significant 2.776 & \significant 1.277 & \significant 1.038 \\
\addlinespace[0.2cm]

MetricX 23 XL  & & \significant 1.423 & \significant 1.423 & \significant 1.705 & \significant 1.781 & \significant 1.051 & \significant 1.916 & \significant 1.214 & \significant 1.802 \\
MetricX 23 XL & \checkmark & \significant 1.395 & \significant 1.353 & \significant 2.166 & \significant 2.059 & \significant 1.300 & \significant 1.881 & \significant 1.726 & 0.955  \\
\addlinespace[0.2cm]

Mistral 7B  &  & \significant 1.494 & \significant 1.220 & \significant 1.296 & \significant 1.285 & \significant 1.222 & \significant 1.208 & \significant 1.018 & \significant 1.030 \\
Mistral 7B V2 &  & \significant 1.665 & \significant 1.317 & \significant 1.357 & \significant 1.324 & \significant 1.311 & \significant 1.235 & \significant 1.035 & \significant 1.094 \\
\addlinespace[0.2cm]

Gemma 2 9B  & & \significant 3.129 & \significant 3.034 & \significant 2.869 & \significant 2.654 & \significant 3.458 & \significant 2.271 & \significant 2.127 & \significant 2.148 \\
Gemma 2 9B V2 &  & \significant 2.972 & \significant 2.935 & \significant 3.029 & \significant 2.508 & \significant 2.861 & \significant 2.286 & \significant 1.959 & \significant 1.806 \\
\addlinespace[0.2cm]

Llama 3.1 70B  &  & \significant 2.126 & \significant 2.597 & \significant 2.559 & \significant 2.397 & \significant 2.387 & \significant 1.844 & \significant 2.238 & \significant 1.447 \\
Llama 3.1 70B V2 &  & \significant 2.195 & \significant 2.791 & \significant 2.520 & \significant 2.279 & \significant 2.386 & \significant 1.855 & \significant 2.253 & \significant 1.507 \\
\addlinespace[0.2cm]

GPT 4o &  & \significant 2.705 & \significant 3.016 & \significant 2.444 & \significant 2.811 & \significant 2.784 & \significant 2.556 & \significant 2.202 & \significant 2.580 \\

\bottomrule
\end{tabular}
% }
\caption{\textbf{Error rate ratio $\Phi(S^F,S^M)$ between gender groups} along with statistical significance tests on individual languages. \colorbox{red!20}{Highlighted values} indicate that error rates for sources with feminine referents are significantly higher than for their masculine counterparts. Unambiguous, Extra-sentential case. Contextual subset of MT-GenEval. $\mathrm{tr}(c)=\checkmark$: context-aware metrics with translated context.} 

\label{app:tab:contextual_non_amb_stat_tests_phi}
\end{table*}

\begin{table*}[!th]
\centering
\small
\begin{tabular}{lcccccccc}
\toprule
\rowcolor{white}
\textbf{Metrics} & \multicolumn{8}{c}{$QE_{GT}(s_F,h_F) / QE_{GT}(s_M,h_M)$ } \\
& de & es & fr & it & pt & ru & hi & ar \\
\midrule
CometKiwi 22 & 0.998  & \significant 0.997 & \significant 0.996 & \significant 0.996 & 0.998  & 0.996  & \significant 0.997 & 0.997  \\
CometKiwi 23 XL & 0.999  & 0.998  & 0.996  & 1.011  & 1.004  & \significant 0.990 & 1.006  & 0.993  \\
CometKiwi 23 XXL & 1.003  & 0.996  & 1.024  & 1.000  & 1.013  & 0.994  & 0.996  & 0.996  \\
\addlinespace[0.2cm]
xCOMET XL & \significant 0.993 & \significant 0.985 & \significant 0.985 & 0.994  & 0.993  & \significant 0.983 & \significant 0.981 & 1.003  \\
xCOMET XXL & 0.999  & 0.995  & 0.994  & 0.999  & 0.999  & 0.991  & 1.005  & 0.996  \\
\addlinespace[0.2cm]

MetricX 23 L & 1.000  & 1.000  & \significant 0.998 & \significant 0.998 & 0.999  & 0.999  & 1.000  & \significant 0.992 \\
MetricX 23 XL & 1.002  & 1.001  & 1.000  & 0.999  & 0.997  & 1.004  & 1.003  & 1.001  \\

\bottomrule
\end{tabular}
\caption{\textbf{Per language  QE score ratios} along with the corresponding statistical significance tests when assessing correctly-inflected feminine and masculine \textit{GT translations}. \colorbox{red!20}{Highlighted values} indicate that correctly-inflected feminine translations receive significantly lower scores compared to their masculine counterparts. Unambiguous, Intra-sentential case. Source instances from counterfactual subset of MTGenEval.}  
\label{app:tab:counterfactual_stat_tests_ratio_QE_GT}
\end{table*}

\begin{table*}[!th]
\centering
\small
\begin{tabular}{lcccccccc}
\toprule
\rowcolor{white}
\textbf{Metrics} & \multicolumn{8}{c}{$\Phi_{GT}(S^F,S^M) = ER_{GT}(S^F) /\ ER_{GT}(S^M)$}  \\
& de & es & fr & it & pt & ru & hi & ar \\
    \midrule
CometKiwi 22 & 0.571  & \significant 1.269 & \significant 2.000 & \significant 1.292 & \significant 1.100 & \significant 2.333 & 0.526  & \significant 1.667 \\
CometKiwi 23 XL & \significant 1.600 & \significant 1.143 & 0.833  & \significant 1.160 & 1.000  & \significant 3.667 & \significant 1.053 & \significant 2.333 \\
CometKiwi 23 XXL & 0.600  & \significant 1.125 & \significant 4.000 & 1.000  & 0.667  & \significant 1.500 & \significant 1.030 & \significant 3.000 \\
\addlinespace[0.2cm]
xCOMET XL & \significant 2.000 & \significant 1.238 & \significant 4.000 & \significant 1.304 & \significant 2.571 & \significant 1.800 & 0.958  & \significant 1.778 \\
xCOMET XXL & 0.667  & \significant 1.286 & \significant 1.750 & \significant 1.038 & \significant 1.429 & 1.000  & \significant 1.167 & \significant 1.500 \\
\addlinespace[0.2cm]
MetricX 23 L & \significant 1.086 & \significant 1.216 & \significant 1.115 & \significant 1.208 & 0.973  & \significant 1.290 & 0.966  & \significant 1.474 \\
MetricX 23 XL & \significant 1.429 & \significant 1.207 & \significant 1.286 & \significant 1.125 & \significant 1.400 & \significant 2.143 & \significant 1.071 & 1.000  \\

\bottomrule
\end{tabular}
\caption{ \textbf{Per language error rate ratios $\Phi_{GT}(S^F,S^M)$ between gender groups,} along with the corresponding statistical significance tests.  \textit{Assessing GT translations}. \colorbox{red!20}{Highlighted values} indicate that error rates for sources with feminine referents are significantly higher than for their masculine counterparts. Unambiguous, Intra-sentential case. Source instances from counterfactual subset of MTGenEval.}
\label{app:tab:counterfactual_stat_tests_ratio_Phi_GT}
\end{table*}

% \begin{table*}[h]
% \centering
% \small
% % \scalebox{0.9}{
% \begin{tabular}{lcccc}
% \toprule
% \textbf{Neural Metrics} & \textbf{Translated Context} & \textbf{Ties Rate(\%)} & \textbf{GEMBA-prompted Metrics} & \textbf{Ties Rate(\%)}  \\
% \midrule
% MetricX 23 L &\checkmark & 0.04 &
% Llama 3.1 70B V1    &          55.45    \\
% MetricX 23 XL &\texttimes & 0.19 & 
% Llama 3.1 70B V2   &          55.01  \\
% MetricX 23 XL &\checkmark & 11.30 &
% Gemma 2 9B V1       &           49.65  \\
% xCOMET XL &\texttimes & 0.25 &
% Gemma 2 9B V2      &      51.67\\
% xCOMET XL &\checkmark & 0.30 &
% Mistral 7B V1         &          85.18\\
% xCOMET XXL &\texttimes & 1.44 &
% Mistral 7B V2      &       82.48  \\
% xCOMET XXL &\checkmark & 1.05 &
% GPT 4o        &     38.34 \\

% \bottomrule
% \end{tabular}
% % }
% \caption{\textbf{Aggregated tie rates} 
%   across all languages for all QE Metrics on the contextual subset of MTGen-Eval. 
% }
% \label{app:tab:ties_contextual_non_amb_llms}
% \end{table*}

\begin{table*}[!t]
\centering
\small
% \scalebox{0.9}{
\subfloat[Neural metrics]{%
    \hspace{.5cm}
    \begin{tabular}{lcc}
    \toprule
    \textbf{Metric} & \textbf{tr(\textit{c})} & \textbf{Tie rate (\%)} \\
    \midrule
    MetricX 23 L &\checkmark & 0.04 \\
    \tabrowsep
    MetricX 23 XL & & 0.19 \\
    MetricX 23 XL &\checkmark & 11.30 \\
    \tabrowsep
    xCOMET XL & & 0.25 \\
    xCOMET XL &\checkmark & 0.30 \\
    \tabrowsep
    xCOMET XXL & & 1.44 \\ 
    xCOMET XXL &\checkmark & 1.05 \\ \bottomrule
    \end{tabular}
} \hspace{1cm}
\subfloat[GEMBA metrics]{%
    \hspace{.5cm}
    \begin{tabular}{lc}
    \toprule
    \textbf{Metric} & \textbf{Tie rate (\%)}  \\
    \midrule
    Llama 3.1 70B V1    &          55.45    \\ 
    Llama 3.1 70B V2   &          55.01  \\
    \tabrowsep
    Gemma 2 9B V1       &           49.65  \\
    Gemma 2 9B V2      &      51.67\\
    \tabrowsep
    Mistral 7B V1         &          85.18\\
    Mistral 7B V2      &       82.48  \\
    \tabrowsep
    GPT 4o        &     38.34 \\
    \bottomrule
    \end{tabular}
}

% }
\caption{\textbf{Aggregated tie rates} 
  across all languages for all QE Metrics on the contextual subset of MT-GenEval with at least one tie. $\mathrm{tr}(c)=\checkmark$: context-aware metrics with translated context.
}
\label{app:tab:ties_contextual_non_amb_llms}
\end{table*}

\begin{table}[!t]
    \centering
    \scalebox{0.8}{
   \begin{tabular}{lrrrr}
   \toprule
  \textbf{Quality}  & \multicolumn{2}{c}{\textbf{\# Stage 1}}  & \textbf{\# Stage 2} \\
  & F & M & \\
  \midrule
  Excellent (50+) & 662 & 654 & 590 \\
  Very Good (40-50) & 282 & 271 & 191\\
  Good (30-40) & 218 & 209 & 140\\
  Fair (20-30) & 194 & 189 & 143\\
  Poor (<20) & 149 & 163 & 114\\
  \midrule
  Total & 1505 & 1486 & 1178\\
\bottomrule
\end{tabular}}
    \caption{Each LP originally has 300 instances. We retain approximately 50\% of the samples.}
    \label{tab:filter_stats}
\end{table}

\end{document}